\documentclass[journal]{IEEEtran}

\usepackage[mathscr]{eucal}
\usepackage[cmex10]{amsmath}
\usepackage{epsfig,epsf,psfrag}
\usepackage{amssymb,amsmath,amsthm,amsfonts,latexsym}
\usepackage{amsmath,graphicx,bm,xcolor,url}
\usepackage[caption=false]{subfig}
\usepackage{fixltx2e}%ordering of single and double column floats
\usepackage{array}%array and tabular environments
\usepackage{verbatim}
\usepackage{bm}
\usepackage{algorithmic, cite}
\usepackage{algorithm}
\usepackage{verbatim}
\usepackage{textcomp}
\usepackage{mathrsfs}
\usepackage{epstopdf}
\catcode`~=11 \def\UrlSpecials{\do\~{\kern -.15em\lower .7ex\hbox{~}\kern .04em}} \catcode`~=13

\allowdisplaybreaks[3]

\newcommand{\calA}{\mathcal{A}}

\newcommand{\calD}{\mathcal{D}}

\newcommand{\calH}{\mathcal{H}}
\newcommand{\calI}{\mathcal{I}}

\newcommand{\calN}{\mathcal{N}}
\newcommand{\calO}{\mathcal{O}}

\newcommand{\calQ}{\mathcal{Q}}

\newcommand{\calU}{\mathcal{U}}

\newcommand{\ba}{\mathbf{a}}
\newcommand{\bA}{\mathbf{A}}
\newcommand{\bb}{\mathbf{b}}
\newcommand{\bB}{\mathbf{B}}
\newcommand{\bc}{\mathbf{c}}
\newcommand{\bC}{\mathbf{C}}
\newcommand{\bd}{\mathbf{d}}
\newcommand{\bD}{\mathbf{D}}
\newcommand{\be}{\mathbf{e}}
\newcommand{\bE}{\mathbf{E}}

\newcommand{\bF}{\mathbf{F}}
\newcommand{\bg}{\mathbf{g}}
\newcommand{\bG}{\mathbf{G}}
\newcommand{\bh}{\mathbf{h}}
\newcommand{\bH}{\mathbf{H}}

\newcommand{\bI}{\mathbf{I}}

\newcommand{\bN}{\mathbf{N}}
\newcommand{\bo}{\mathbf{o}}

\newcommand{\bp}{\mathbf{p}}

\newcommand{\bq}{\mathbf{q}}
\newcommand{\bQ}{\mathbf{Q}}

\newcommand{\bR}{\mathbf{R}}
\newcommand{\bs}{\mathbf{s}}

\newcommand{\bT}{\mathbf{T}}
\newcommand{\bu}{\mathbf{u}}
\newcommand{\bU}{\mathbf{U}}
\newcommand{\bv}{\mathbf{v}}

\newcommand{\bw}{\mathbf{w}}
\newcommand{\bW}{\mathbf{W}}
\newcommand{\bx}{\mathbf{x}}

\newcommand{\bY}{\mathbf{Y}}
\newcommand{\bz}{\mathbf{z}}

\DeclareMathAlphabet{\mathbsf}{OT1}{cmss}{bx}{n}
\DeclareMathAlphabet{\mathssf}{OT1}{cmss}{m}{sl}% slanted sans serif

\DeclareSymbolFont{bsfletters}{OT1}{cmss}{bx}{n}
\DeclareSymbolFont{ssfletters}{OT1}{cmss}{m}{n}
\DeclareMathSymbol{\bsfGamma}{0}{bsfletters}{'000}
\DeclareMathSymbol{\ssfGamma}{0}{ssfletters}{'000}
\DeclareMathSymbol{\bsfDelta}{0}{bsfletters}{'001}
\DeclareMathSymbol{\ssfDelta}{0}{ssfletters}{'001}
\DeclareMathSymbol{\bsfTheta}{0}{bsfletters}{'002}
\DeclareMathSymbol{\ssfTheta}{0}{ssfletters}{'002}
\DeclareMathSymbol{\bsfLambda}{0}{bsfletters}{'003}
\DeclareMathSymbol{\ssfLambda}{0}{ssfletters}{'003}
\DeclareMathSymbol{\bsfXi}{0}{bsfletters}{'004}
\DeclareMathSymbol{\ssfXi}{0}{ssfletters}{'004}
\DeclareMathSymbol{\bsfPi}{0}{bsfletters}{'005}
\DeclareMathSymbol{\ssfPi}{0}{ssfletters}{'005}
\DeclareMathSymbol{\bsfSigma}{0}{bsfletters}{'006}
\DeclareMathSymbol{\ssfSigma}{0}{ssfletters}{'006}
\DeclareMathSymbol{\bsfUpsilon}{0}{bsfletters}{'007}
\DeclareMathSymbol{\ssfUpsilon}{0}{ssfletters}{'007}
\DeclareMathSymbol{\bsfPhi}{0}{bsfletters}{'010}
\DeclareMathSymbol{\ssfPhi}{0}{ssfletters}{'010}
\DeclareMathSymbol{\bsfPsi}{0}{bsfletters}{'011}
\DeclareMathSymbol{\ssfPsi}{0}{ssfletters}{'011}
\DeclareMathSymbol{\bsfOmega}{0}{bsfletters}{'012}
\DeclareMathSymbol{\ssfOmega}{0}{ssfletters}{'012}

\DeclareMathOperator*{\argmax}{arg\,max}
\DeclareMathOperator*{\argmin}{arg\,min}

\DeclareMathOperator{\sgn}{sgn}

\newtheorem{theorem}{Theorem}
\newtheorem{lemma}[theorem]{Lemma}

\newtheorem{definition}{Definition}

\newtheorem{remark}{Remark}

\newtheorem{assumption}{Assumption}
\newtheorem{data model}{Data Model}

\newcommand{\qednew}{\nobreak \ifvmode \relax \else
      \ifdim\lastskip<1.5em \hskip-\lastskip
      \hskip1.5em plus0em minus0.5em \fi \nobreak
      \vrule height0.75em width0.5em depth0.25em\fi}

\begin{document}
\title{Outlier Detection and Data Clustering via Innovation Search}

\author{\textbf{Mostafa~Rahmani and Ping Li}\\
Cognitive Computing Lab\\
Baidu Research\\
10900 NE 8th St. Bellevue, WA 98004, USA\\
\{mostafarahmani, liping11\}@baidu.com
}

\markboth{}%
%\markboth{Journal of \LaTeX\ Class Files,~Vol.~11, No.~4, %December~2012}%
{Shell \MakeLowercase{\textit{et al.}}: Bare Demo of IEEEtran.cls for Journals}
\maketitle

\begin{abstract}
The idea of Innovation Search was proposed as a data clustering method in which the directions of innovation were utilized to compute the adjacency matrix and it was shown that Innovation Pursuit can notably outperform the self representation based subspace clustering methods. In this paper, we present a new discovery that the directions of innovation can be used to design a provable and  strong  robust (to outlier) PCA method.
The proposed approach, dubbed iSearch, uses the direction search optimization problem to compute an optimal direction corresponding to each data point.
An outlier by definition is a data point which
does not participate in forming a low dimensional structure
with a large number of data points in the data.
In other word, an outlier carries  innovation with respect to most of the other data points.
iSearch utilizes the directions of innovation to measure the innovation of the data points and it identifies the outliers as the most innovative data points.
Analytical performance guarantees are derived for the proposed robust PCA method  under different models for the distribution of the outliers including randomly distributed outliers, clustered outliers, and linearly dependent outliers.
In addition,
we study the problem of outlier detection in a union of subspaces and
it is shown that iSearch provably recovers the span of the inliers when the inliers lie in a union of subspaces.  Moreover, we present theoretical studies which show that  the proposed measure of innovation remains stable in the presence of noise and the performance of iSearch is robust to noisy data.
In the challenging scenarios in which the outliers are close to each other or they are close to the span of the inliers, iSearch is  shown to remarkably outperform most of the existing methods. The presented method shows that the directions of innovation~\cite{rahmani2015innovation,rahmani2017subspacedi} are useful representation of the data which can be used to perform both data clustering and outlier detection.

%The proposed approach is one of the  few robust PCA methods which is provably robust to the randomly distributed outliers, the clustered outliers, and the linearly dependent outliers.
\end{abstract}


\IEEEpeerreviewmaketitle

\section{Introduction}
Principal Component Analysis (PCA) has been extensively
used to reduce dimensionality by finding linear projections
of high-dimensional data into lower dimensional subspaces.
 PCA finds an $r$-dimensional linear subspace by solving
\begin{eqnarray}
\underset{ \hat{\bU}}{\min} \: \:  \| \bD - \hat{\bU} \hat{\bU}^T \bD \|_F^2 \quad \text{subject to} \quad \hat{\bU}^T \hat{\bU} = \bI \:,
\label{eq:aval}
\end{eqnarray}
where $\bD \in \mathbb{R}^{M_1 \times M_2}$ is the given data, $\hat{\bU}$ is an orthonormal basis for the $r$-dimensional subspace, $\bI$ is the identity matrix, $\| \cdot \|_F$ denotes
the Frobenius norm, $M_2$ is the number of data points, and $M_1$ is the dimensionality of the ambient space.
While PCA is useful when the data has
low intrinsic dimension, it is  sensitive to outliers
in the sense that the solution to (\ref{eq:aval}) can arbitrarily deviate from
the true underlying subspace if a small portion of the data is
not contained in this low-dimensional subspace.

%Outlier detection is an important problem in unsupervised learning in which the outlier detector should find the  data points which do not follow the underlying pattern of the data. In other word, the outlier detection algorithm infers the underlying structure of the data and identifies the outliers based on their innovation with respect to the dominant structure of the data.
In addition to the sensitivity of PCA to outliers,
detecting the outlying data points is also an important research problem in unsupervised machine learning.
%h rare
%patterns or events of interest, such as important regions in an
%image [33], malignant tissues [34], or web attacks [35].
Outliers are associated with important rare events such as malignant tissues~\cite{karrila2011comparison}, the failures of a system~\cite{hauskrecht2013outlier,harada2017log,lamport11}, web attacks~\cite{kruegel2003anomaly}, and misclassified data points~\cite{rahmani2017coherence22f,gitlin2018improving}. %In some applications, the outliers can dominate the data [].
%For instance, in [] we showed that the problem of linear regression with permuted measurements
%can be translated into an outlier detection problem. In this problem,
%an unknown permutation can permute most of the measurements. This is equivalent to an outlier detection problem in which most of the data points are outliers.
% For instance, in the robust (to outlier) PCA problem the column space of the low rank componetn can be recovered as the span of a small subset of the inliers.
In this paper,
the proposed outlier detection method is introduced as a robust Principal Component Analysis (PCA) algorithm, i.e., the inliers lie in a low dimensional subspace. %PCA is arguably the most widely used data analysis tool for dimension reduction. However,  PCA does not consider the presence of the outliers that prevail much of the real world data.  The output of PCA can significantly deviate from the correct subspace even if a small portion of the data is affected by the outliers.

In the literature of robust PCA, two main models for the data corruption are considered: the element-wise data corruption model and the column-wise data corruption model. These two models are corresponding to two completely different robust PCA problems. In the element-wise model, it is assumed that a small subset of the elements of the data matrix are corrupted and the \textcolor{black}{support of the corrupted elements is random}. Thus, the corrupted elements are not concentrated in any column/row of the data. This problem is known as the low rank plus sparse matrix decomposition problem~\cite{lamport22,lamport1,liu2014recovery,liu2016low}.
%In the second model, a random subset of the elements are corrupted. With the second model, the robust PCA problem is equivalent to the low rank plus sparse matrix decomposition problem studied in [].
In the column-wise model,  a subset of the columns of the data  \textcolor{black}{are affected by the data corruption}~\cite{lamport10,zhang2014novel,lerman2014fast,fischler1981random,li1985projection,choulakian2006l1,feng2012robust,mccoy2011two,hardt2013algorithms,zhang2016robust,you2017provable,markopoulos2014optimal}.
We solely  focus on the column-wise data corruption model. Specifically, it is assumed that the given data matrix $\bD$ follows Data Model 1.
\begin{data model}
The matrix $\bD$ can be expressed as
$
\bD =  [\bB \hspace{.2cm} \bA] \: \bT \:,
$
where $\bA \in \mathbb{R}^{M_1 \times n_i}$, $\bB \in \mathbb{R}^{M_1 \times n_o}$,  $\bT$ is an arbitrary permutation matrix, and $[\bB \hspace{.2cm} \bA]$ represents the concatenation of matrices $\bA$ and $\bB$.
 The columns of $\bA$ lie in an $r$-dimensional subspace $\calU$. The columns of $\bB$ do not lie entirely in $\calU$, i.e., the $n_i$ columns of $\bA$ are the inliers and the $n_o$ columns of $\bB$ are the outliers. The orthonormal matrix $\bU \in \mathbb{R}^{M_1 \times r}$ is a basis for $\calU$.
\end{data model}

\noindent
%In the robust PCA problem, the main task is to recover $\calU$.
The output of a robust PCA method is a basis for $\calU$.
If  $\calU$ is estimated accurately, the outliers can be  located using a simple subspace projection~\cite{li2015identifying}.

\textbf{Summary of Contributions:}
 The main contributions can be summarized as follows.
\begin{itemize}\vspace{-0.08in}
\item iSearch introduces a new idea to the robust PCA problem. iSearch utilizes the directions of innovation to measure the Innovation of the data points.  It is shown that the proposed approach mostly outperforms the exiting methods in handling close outliers and noisy data.
\vspace{-0.03in}
\item  The proposed approach and the CoP method presented in~\cite{rahmani2017coherence22f}, to the best of our knowledge, are the only robust PCA methods which are supported with analytical performance guarantees
    under different models for the distributions of the outliers including
    the  randomly distributed outliers, the clustered outliers, and the linearly dependent outliers.
\vspace{-0.03in}
\item In addition to considering several models for the distribution of the outliers,  we provide analytical  performance guarantees  under different models  for  the  distribution  of  the  inliers too. The presumed models include the union of subspaces and the uniformly at random distribution on $\calU \cap \mathbb{S}^{M_1 -1}$ where  $\mathbb{S}^{M_1-1}$ denotes the unit $\ell_2$-norm sphere in $\mathbb{R}^{M_1}$. Moreover, the stability of iSearch in the presence of additive noisy is studied and it is proved that iSearch is robust to noisy data.
\end{itemize}\vspace{-0.05in}

%        $\bullet$ \textcolor{black}{ The proposed approach shows that Innovation Search can perform data clustering and outlier detection simultaneously.  }

\subsection{Notation}
Given a matrix $\bA$, $\| \bA \|$ denotes its spectral norm. For a vector $\ba$, $\| \ba \|_p$ denotes its $\ell_p$-norm and $\ba(i)$ its $i^{\text{th}}$ element. Given two matrices $\bA_1$ and $\bA_2$ with an equal number of rows, the matrix
$
\bA_3 = [\bA_1 \: \: \bA_2]
$
is the matrix formed by concatenating their columns. For a matrix $\bA$, $\ba_i$ denotes its $i^{\text{th}}$ column.
%The function orth$(\cdot)$ % is defined similar to the function orth$(\cdot)$ in MATLAB, which
%returns an orthonormal basis for the column-space of its matrix argument.
The subspace $\calU^{\perp}$ is the complement of $\calU$. The cardinality of set $\calI$ is defined as  $| \calI |$. Also, for any positive integer n, the index
set $\{1,...,n\}$ is denoted $[n]$. The coherence between vector $\ba$ and subspace $\calH$ with orthonormal basis $\bH$ is defined as $\| \ba^T \bH \|_2$.

\section{Related Works}
In this section, we  review some of the important related works in robust (to outlier) PCA and the Innovation Pursuit method. We refer the reader to~\cite{lerman2018overview,rahmani2017coherence22f} for a more comprehensive review on the topic.
Some of the earliest approaches to robust PCA relied on
robust estimation of the data covariance matrix, such as the minimum covariance determinant, the minimum volume ellipsoid, and the Stahel-Donoho estimator~\cite{lamport47}.
They mostly compute a full SVD
or eigenvalue decomposition in each iteration and generally
have no explicit performance guarantees. The performance of
these approaches greatly degrades when $\frac{n_i}{n_o} < 0.5$.
To enhance the robustness to the presence of the outliers, another approach is to replace  the Frobenius Norm in the cost function of PCA  with  $\ell_1$-norm~\cite{lamport18,lamport29} because $\ell_1$-norm were shown to be robust to the presence of the outliers~\cite{decod,lamport29}. However, $\ell_1$-norm does not leverage the column-wise structure of the corruption matrix.  The authors of~\cite{lamport21} modified the $\ell_1$-norm minimization problem used in~\cite{lamport29} and replaced it with an $\ell_{1,2}$-norm minimization problem. In~\cite{lerman2015robustnn} and~\cite{zhang2014novel}, the optimization problem used in~\cite{lamport21} was relaxed to two different convex optimization problems. The authors of~\cite{lerman2015robustnn} /~\cite{zhang2014novel} provided sufficient conditions \textcolor{black}{under which} the optimal point of the convex optimization problem proposed in~\cite{lerman2015robustnn} /~\cite{zhang2014novel} is guaranteed to be equal to a projection matrix whose column-space is equal to $\calU$ / $\calU^{\perp}$.
The approach presented in~\cite{tsakiris2015dual} focused on the scenario in which the data is predominantly unstructured outliers and the number of outliers is larger than $M_1$.
In~\cite{tsakiris2015dual}, it is essential to assume that the outliers are randomly distributed on $\mathbb{S}^{S - 1}$ and the inliers are distributed randomly on the intersection of $\mathbb{S}^{S - 1}$ and $\calU$.
%It was shown that the optimal point of the linear optimization problem used in~\cite{tsakiris2015dual} can be used to reject a subset of the outliers. %The strong assumptions used in~\cite{tsakiris2015dual} makes this approach inapplicable to the applications in which the inliers or the outliers are structured.
In~\cite{lamport10}, a convex optimization problem was proposed which decomposes the data into a low rank component and a column sparse component. The approach presented in~\cite{lamport10} is provable but it requires $n_o$ to be significantly smaller than $n_i$.
The outlier detection method proposed in~\cite{soltanolkotabi2012geometric} assumes that the outliers are randomly distributed on $\mathbb{S}^{M_1 - 1}$ and a small number of them are not linearly dependent which means that~\cite{soltanolkotabi2012geometric} is not able to detect the linearly dependent outliers and the outliers which are close to each other.

\subsection{Review of Innovation Pursuit for Data Clustering}
In contrast to most of the  spectral clustering based subspace clustering methods which utilize self-representation to compute the affinity matrix, the authors of~\cite{rahmani2015innovation,rahmani2017subspacedi} proposed Innovation Pursuit which utilized the directions of innovation to compute the affinity matrix. A convex optimization problem was introduced which finds an optimal direction  corresponding to each data point. The Direction of Innovation corresponding to data point $\bd$ is closely aligned with $\bd$ but it has the minimum projection on the other data points. In~\cite{rahmani2015innovation,rahmani2017subspacedi}, it was shown that Innovation Pursuit can significantly outperform the self representation based methods (e.g. Sparse Subspace Clustering~\cite{lamport7}) specifically in the scenarios in which the subspaces intersect.  In this paper, we present a new discovery  that the Directions of Innovation can also be utilized to devise a novel, strong, and provable robust PCA method. The table of Algorithm 1 shows the procedure to use the directions of innovation for both data clustering and outlier detection.

\subsection{Connection and Contrast to Coherence Pursuit}
In~\cite{rahmani2017coherence22f},  the Coherence Pursuit (CoP) method was proposed as an outlier detection method. CoP computes  the Coherence Values for all the data points to rank the data points.
The Coherence value corresponding to data column $\bd$ is a measure of resemblance between $\bd$  and the rest of the data columns. %Subsequently, CoP recovers the  column-space of $\bA$ using a subset of the columns which are corresponding to the largest Coherence Values.
CoP uses the inner product between $\bd$ and the rest of the data points to measure the resemblance between $\bd$ and the rest of data. In sharp contrast, iSearch leverages the directions of innovation  to measure the innovation of each data point and ranks the data points accordingly. We show through theoretical studies and numerical experiments that finding the optimal directions makes iSearch notably stronger than CoP in detecting outliers which carry weak innovation.

%\subsection{Connection and Contrast to Innovation Pursuit}

%In~\cite{rahmani2015innovation},  Innovation Pursuit was proposed as a new subspace clustering method. The optimization problem proposed in~\cite{rahmani2015innovation} finds a direction in the span of the data such that it is orthogonal to the maximum number of data points. In addition, the idea of Innovation Pursuit was extended into a new spectral clustering based subspace segmentation method, dubbed as Direction Search based Subspace Clustering (DSC)~\cite{rahmani2015innovation}. In this paper, we present a new discovery about the applications of  Innovation Pursuit. It is shown that the idea of innovation search can be used to design a new, provable, and strong outlier detection algorithm. In the proposed approach, an optimization problem similar to the linear optimization problem  in~\cite{rahmani2015innovation} is used to measure the innovation of the data points.

\begin{algorithm}
\caption{Robust Subspace Recovery (Outlier Detection) and Data Clustering Using the Directions of Innovation }
{
\textbf{Input.} The input is data matrix $\bD \in \mathbb{R}^{M_1 \times M_2}$.

\smallbreak
\textbf{1. Data Preprocessing.}  \\
\textbf{1.1} Define $\bQ \in \mathbb{R}^{M_1 \times r_d}$ as the matrix of first $r_d$ left singular vectors of $\bD$ where $r_d$ is the number of non-zero singular values. Set $\bD = \bQ^T \bD$. If dimensionality reduction is not required, skip this step.  \\
\textbf{1.2} Normalize the $\ell_2$-norm of the columns of $\bD$, i.e., set $\bd_i$ equal to $\bd_i / \| \bd_i \|_2$ for all $1 \le i \le M_2$.

\smallbreak
\textbf{2. Direction Search.}
Define $\bC^{*} \in \mathbb{R}^{r_d \times M_2}$ such that $\bc_i^{*} \in \mathbb{R}^{r_d \times 1}$ is the optimal point of
$$
\underset{ \bc}{\min} \: \:  \| \bc^T \bD \|_1 \quad \text{subject to} \qquad \bc^T \bd_i = 1
$$
or define $\bC^{*} \in \mathbb{R}^{r_d \times M_2}$  as the optimal point of
\begin{eqnarray}
\underset{ \bC}{\min} \: \:  \| \bC^T \bD \|_{1} \quad \text{subject to} \qquad \text{diag}(\bC^T \bD) = \textbf{1} \:.
\label{opt:kolli}
\end{eqnarray}

\smallbreak
\textbf{\textcolor{blue}{3. If the Task is Robust Subspace Recovery (Outlier Detection):}}
\smallbreak
\textbf{3.1 Computing the Innovation Values. } Define vector $\bx \in \mathbb{R}^{M_2 \times 1}$ such that $\bx(i) = 1 / \| \bD^T \bc_i^{*}  \|_1$.
% C:\Users\ccl\Desktop\Mostafa\iSearch_files\iSearch Figs

\smallbreak
\textbf{3.2 Building Basis. } Construct matrix $\bY$ from the
columns of $\bD$ corresponding to the smallest elements of $\bx$ such that they span an r-dimensional subspace.

\smallbreak
\textbf{3.3 Output:} The column-space of $\bY$ is the identified subspace.

\smallbreak
\textbf{\textcolor{blue}{3. If the Task is Data Clustering:}}
\smallbreak
\textbf{3.1} Define $\bW = \big| \bC^T \bD \big|$.

\textbf{3.2} Apply spectral clustering to $\bW + \bW^T$ (in some applications, pre-processing matrix $\bW$ according to the procedure described  in~\cite{rahmani2017subspacedi} can improve the performance).

\textbf{3.3 Output:} The identified clusters.

 }
\end{algorithm}

\section{Proposed Approach}

Algorithm 1
presents the proposed outlier detection method along with the definition of the used symbols.
In order to show the connection of proposed robust PCA method with the Innovation Search based clustering method~\cite{rahmani2017subspacedi,rahmani2015innovation}, Table 1 demonstrates both algorithms.
iSearch consists of three steps. In the next subsections, Step 2 and Step 3 are discussed. In this paper, we use an ADMM solver to solve (\ref{opt:kolli}). The computation complexity of the solver is
$\calO(\max(M_1 M_2^2 ,  M_1^2 M_2))$. If PCA is used  in the prepossessing step to reduce the  dimensionality of the data to $r_d$, the computation complexity of the solver is
$\calO(\max(r_d M_2^2 , r_d^2 M_2))$ \footnote{If the data is noisy, $r_d$ should be set equal to the number of dominant singular values.  In the numerical experiments, we set $r_d$ equal to the index of the largest singular value which is less than or equal to 0.01 $\%$ of the first singular value. }.

\subsection{An Illustrative Example for Innovation Value}
\label{sect:illus}
A synthetic numerical example is presented to explain the main idea behind iSearch.
Assume $\bD \in \mathbb{R}^{40 \times 250}$, $n_i = 200$, $n_o=50$, $r=5$ and suppose that $\bD$ follows Assumption 1.
\begin{assumption}
 % the $m$-dimensional ambient space.
The columns of $\bA$ are drawn  uniformly at random from $\calU \cap \mathbb{S}^{M_1 -1}$. The columns of $\bB$ are drawn  uniformly at random from $\mathbb{S}^{M_1 - 1}$. To simplify the exposition and notation, it is assumed without loss of generality that $\bT$ in Data Model 1 is the identity matrix, i.e, $\bD = [\bB \: \: \bA]$.
\label{assum_DistUni}
\end{assumption}

\begin{figure}[h!]
\begin{center}
\mbox{\hspace{-0.15in}
\includegraphics[width=1.95in]{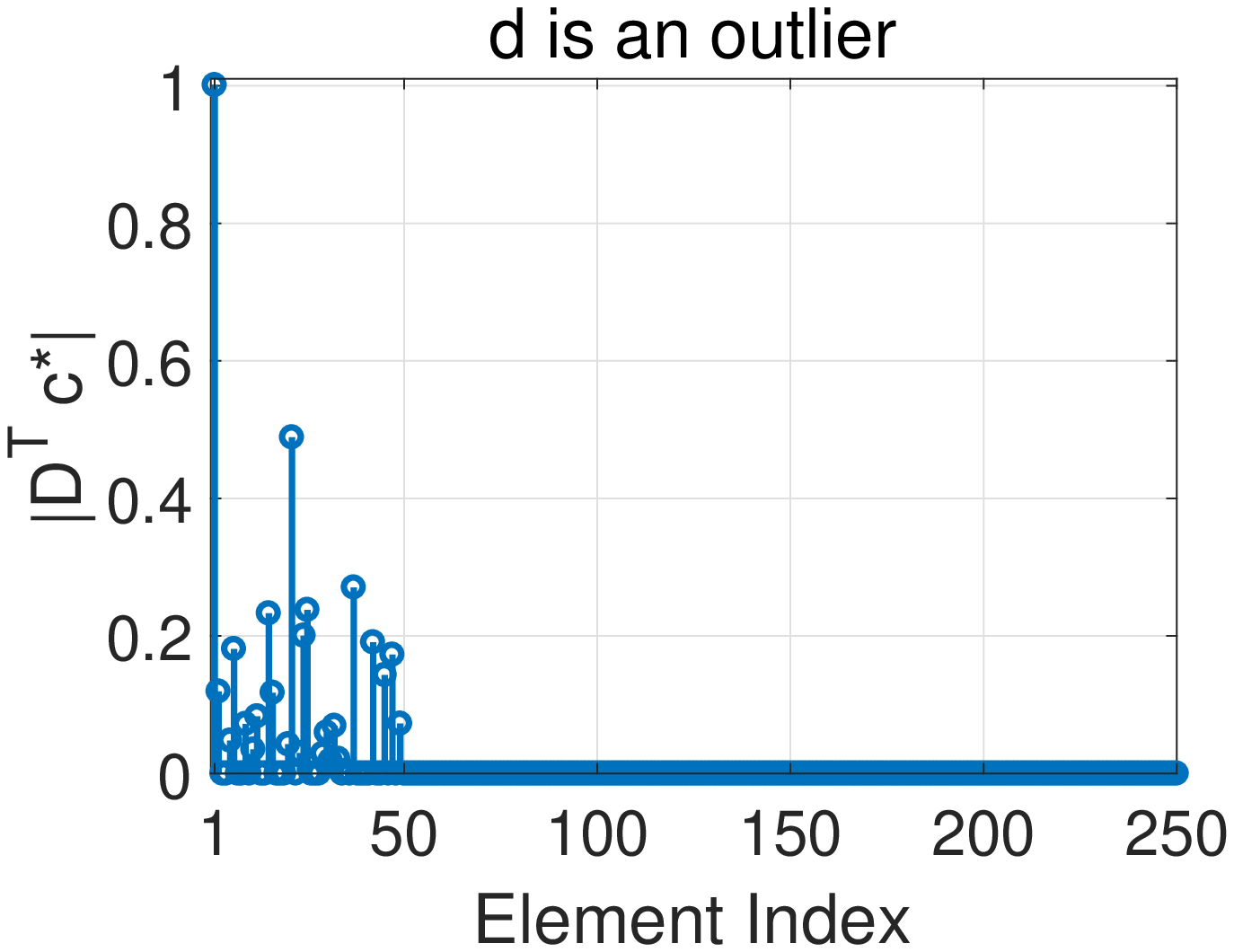}\hspace{-0.15in}
\includegraphics[width=1.95in]{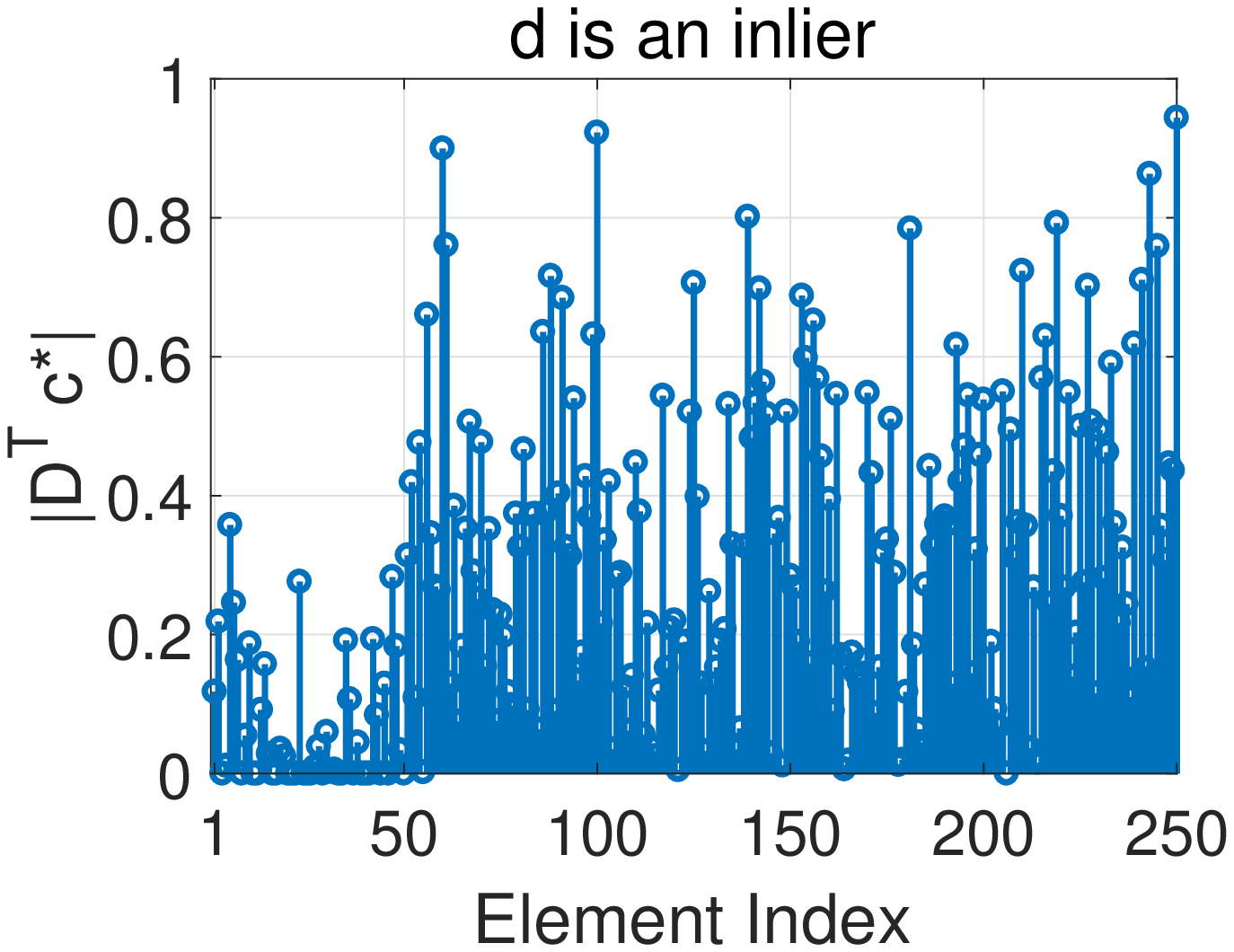}
}
\mbox{\hspace{-0.15in}
\includegraphics[width=1.95in]{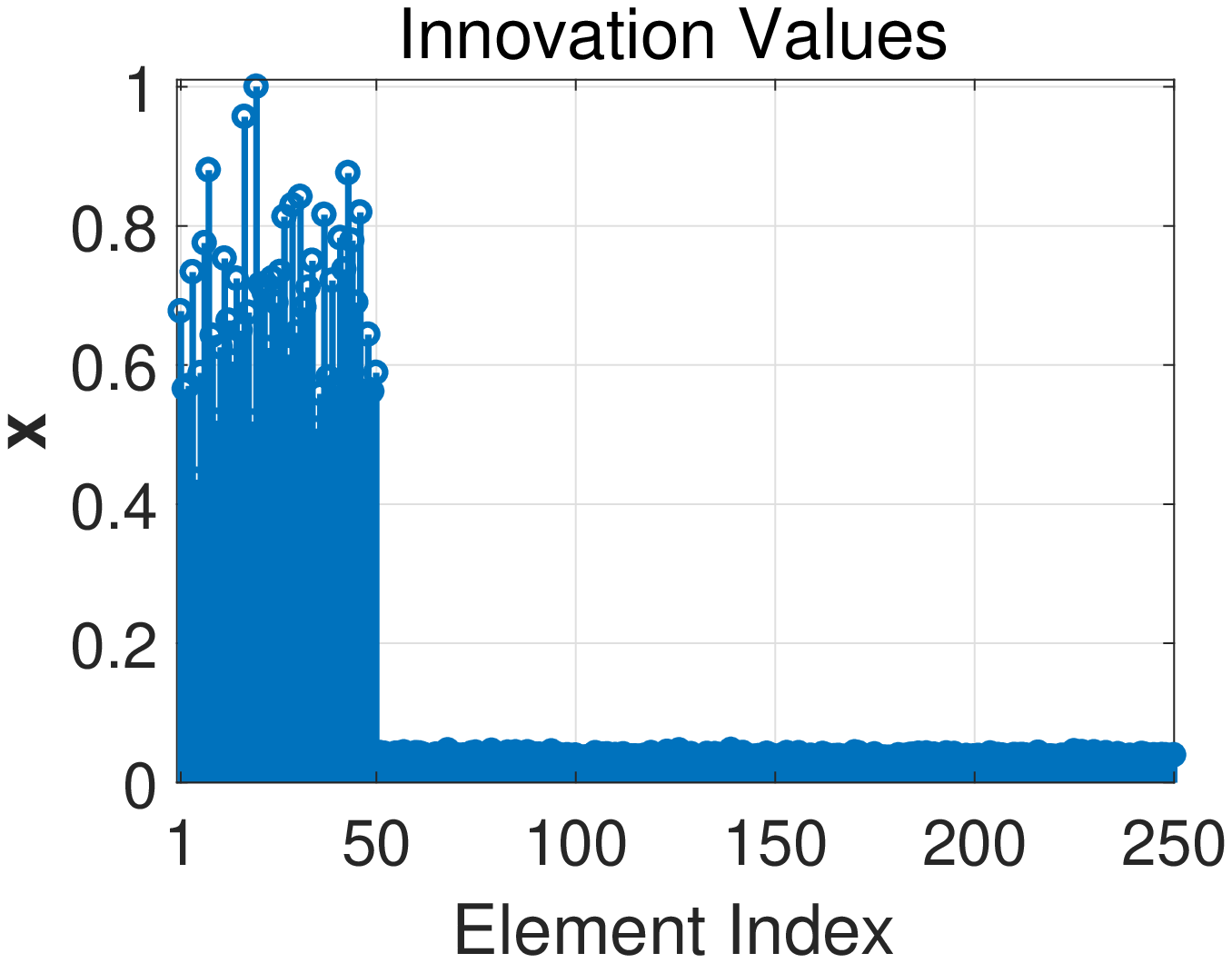}\hspace{-0.15in}
\includegraphics[width=1.95in]{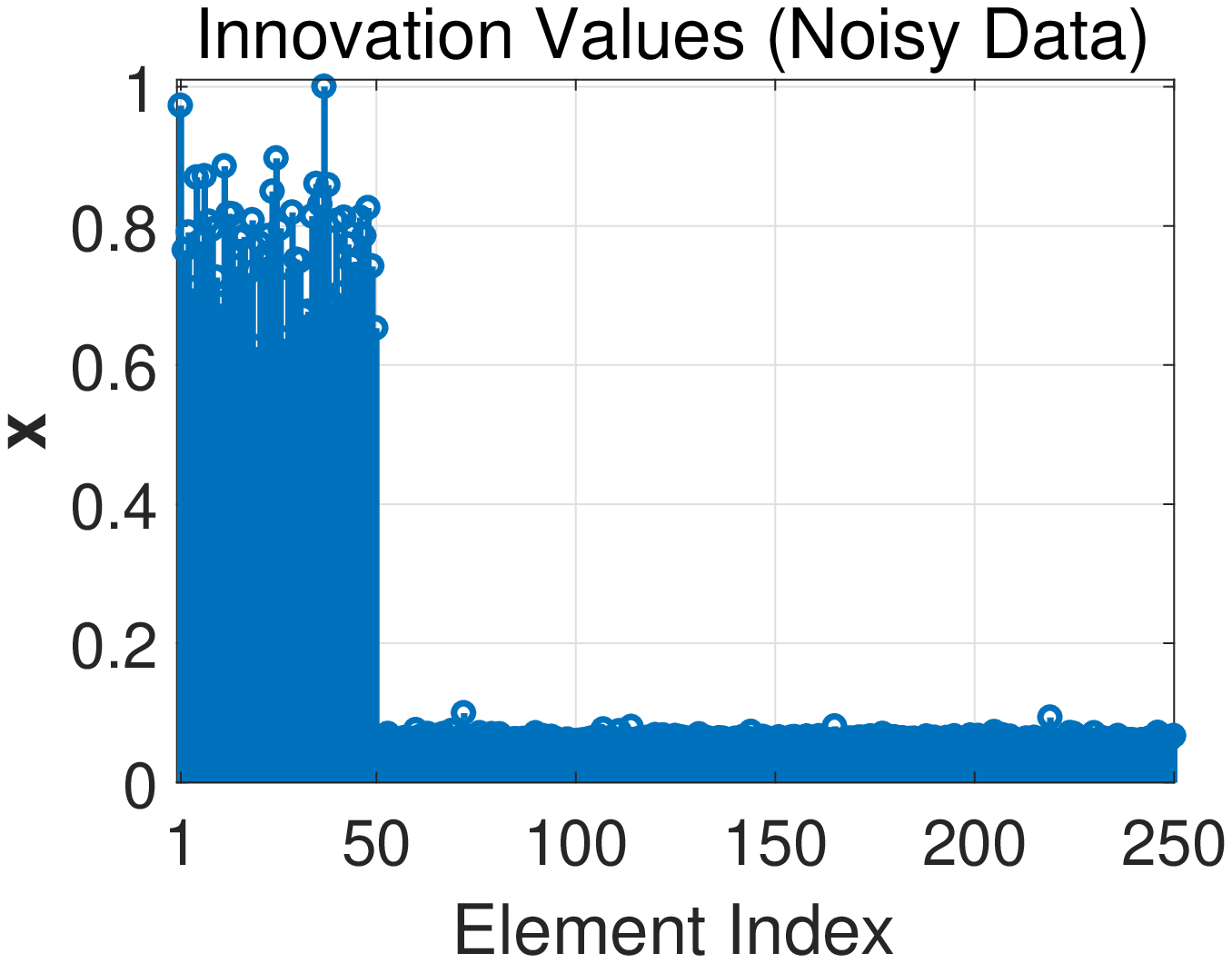}
}
\end{center}
\vspace{-0.15in}
           \caption{In this example, the first 50 columns are outliers. The upper left  panel shows vector $ | \bD^T \bc^{*} |$ when $\bd$ is an outlier.  The upper right  panel depicts $| \bD^T \bc^{*} |$ when $\bd$ is an inlier. The bottom left plot shows the Innovation Values corresponding to all the data points (vector $\bx$ was defined in Algorithm 1). The bottom right  plot also shows   the Innovation Values when the inliers are noisy ($\| \bA \|_F^2/ \|\bE\|_F^2 = 4$ where $\bE \in \mathbb{R}^{M_1 \times n_i}$ denotes the additive noise).}
    \label{fig:show1}
\end{figure}

\noindent
Define $\bd$ as a column of $\bD$ and define $\bc^{*}$ as the optimal point of
\begin{eqnarray}
\underset{ \bc}{\min} \: \:  \| \bc^T \bD \|_1 \quad \text{subject to} \qquad \bc^T \bd = 1 \:,
\label{opt:asli1}
\end{eqnarray}
and define the Innovation Value corresponding to $\bd$ as $1/\|  \bD^T \bc^{*} \|_1$.
The main idea of iSearch is that $\bc^{*}$ shows two completely different behaviours with respect to $\calU$ (when $\bd$ is an outlier and when $\bd$ is an inlier). First assume that $\bd$ is an outlier.
The  optimization problem (\ref{opt:asli1}) searches for a direction whose projection on $\bd$ is non-zero and it has the minimum projection on the rest of the data points.
Since $\bd$ is an outlier, $\bd$ has a non-zero projection on $\calU^{\perp}$. \textcolor{black}{ In addition, since $n_i$ is large, (\ref{opt:asli1}) searches for a direction in the ambient whose projection on $\calU$ is as weak as possible.} Therefore, $\bc^{*}$ lies in $\calU^{\perp}$ or it is close to $\calU^{\perp}$.
The left plot of Fig.~\ref{fig:show1} shows $\bD^T \bc^{*}$ when $\bd$ is an outlier. In this case, $\bc^{*}$ is orthogonal to all the inliers.
Accordingly, when $\bd$ is an outliers, $\| \bD^T \bc^{*} \|_1$ is approximately equal to $\| \bB^T \bc^{*} \|_1$. On the other hand, when $\bd$ is an inlier, the linear constraint strongly discourages $\bc^{*}$ to lie in $\calU^{\perp}$ or to be close to $\calU^{\perp}$.
Inliers lie in a low dimensional subspace and mostly they are  close to each other.
Since $\bc^{*}$ has a strong projection on $\bd$, \textcolor{black}{it has strong projections on many of the inliers}. Accordingly, the value of $\| \bA^T \bc^{*} \|_1$ is much larger when $\bd$ is an inlier.
Therefore, the Innovation Value corresponding to an inlier is smaller than the  Innovation Value corresponding to an outlier because $\|\bA^T \bc^{*} \|_1$ is much larger when $\bd$ is an inliers.
Fig.~\ref{fig:show1} compares the vector $ \bD^T \bc^{*} $ when $\bd$ is an outliers with the same vector when $\bd$ is an inlier. In addition, it shows the vector of Innovation Values (right plot). One can observe that the Innovation Values make the outliers clearly distinguishable.

%  ins an outlierd, p=1

\subsection{Building the Basis Matrix}

The data points corresponding to the least Innovation Values are used to construct the basis matrix $\bY$. If the data follows Assumption~\ref{assum_DistUni}, the $r$ data points corresponding to the $r$ smallest Innovation Values span $\calU$ with overwhelming probability~\cite{vershynin2010introduction}.
In practise, the algorithm should continue adding new columns to $\bY$ until the columns of  $\bY$ spans an $r$-dimensional subspace. This approach requires to check the singular values of $\bY$  several times. We propose two techniques to  \textcolor{black}{avoid this extra steps}. The first approach is based on the side information that we mostly have about the data.  In many applications, we can have an upper-bound on $n_o$ because outliers are mostly associated with rare events. If we know that the number of outliers is less than $y$ percent of the data,  matrix $\bY$ can be constructed using $(1 - y)$ percent of the data columns which are corresponding to the least Innovation Values. The second approach is the adaptive column sampling method proposed in~\cite{rahmani2017coherence22f}. The adaptive column sampling method avoids sampling redundant columns.
%via projecting the data columns into the complement of the span of the sampled columns.
%In the presence of additive noise, the adaptive sampling method is run multiple times to ensure that the dominant singular values of $\bY$ are sufficiently larger than the singular values which are corresponding to the noise component.
%\textcolor{blue}{ The details of the adaptive sampling method is provided in the attached extended version of this paper. }

\subsection{Robustness to noise}
The performance of iSearch is  robust to the presence of noise because even if the inliers are noisy, the optimal direction of (\ref{opt:asli1}) corresponding to an outlier remains incoherent with $\calU$. The reason is that if the optimal direction is not incoherent with $\calU$ and the inliers are not dominated by noise, $\|  \bA^T \bc^{*} \|_1$ is large because the inliers lie in a low dimensional subspace and they are close to each other. In Section~\ref{sec:nooise}, we provide a guarantee for the performance of iSearch with noisy data, an upper-bound for  $\| \bU^T \bc^{*} \|_2$ is established, and it is shown that the value of the upper-bound is proportional to $\sqrt{r/ (M_1 n_i^2) }$. The right plot in Fig.~\ref{fig:show1} shows the innovation values for the example described in Section~\ref{sect:illus}. In this plot, the inliers are noisy, $\bD = [\bB \:\:\: (\bA + \bE)]$ where $\bE$ denotes the additive noise, and $\| \bA\|_F^2/ \| \bE \|_F^2 = 4$. One can observe that Innovation Values  clearly distinguishes the outliers from the inliers.

\section{Theoretical Studies}
\label{sec:theo}

In this section, we analyze the performance of the proposed approach with three different models for the distribution of the outliers: unstructured outliers, clustered outliers, and linearly dependent outliers. Moreover, we analyze iSearch with two different models for the distribution of the inliers. These models include the union of subspaces and uniformly at random distribution on $\calU \cap \mathbb{S}^{M_1 - 1}$. In addition,
the performance of iSearch in the presence of noise is studied and it is shown that the direction of innovation is a reliable tool  when the data is noisy.
 The theoretical results are followed by short discussions which highlight the important aspects of the theorems and a short description of the proof. The proofs of all the presented theoretical results are included in Appendix.  %In all the theorems, $\bQ = \bI$ where $\bI \in \mathbb{R}^{M_1 \times M_1}$ is the identity matrix.

\subsection{Randomly Distributed Outliers} \label{sec:theory_random}
In this part, it is assumed that $\bD$ follows Assumption~\ref{assum_DistUni}.
In order to guarantee the performance of the proposed approach, it is enough to show that the Innovation Values corresponding to the outliers are greater than the Innovation Values corresponding to the inliers. In other word, it suffices to show
\begin{eqnarray}
%\begin{aligned}
&\max \left(  \{ 1/\| \bD^T \bc_i^{*} \|_1 \}_{i=n_o+1}^{M_2}  \right) <
  \min \left(  \{ 1/ \| \bD^T \bc_j^{*} \|_1 \}_{j=1}^{n_o}  \right) \:.
%\end{aligned}
\label{cond:Innovation_condition}
\end{eqnarray}
Before we state the theorem, let us provide the following definitions and remarks.
\begin{definition}
Define $\bc_j^{*} = \underset{  \bd_j^T\bc = 1 }{\arg\min} \: \:  \| \bc^T \bD \|_1$. In addition, define   $\chi = \max \big( \{ \| \bc_j^{*} \|_2 \}_{i=1}^{n_o} \big)$, $\chi^{'} = \max \big( \{ \| \bc_j^{*} \|_2 \}_{i=n_o+1}^{M_2} \big)$,  and $n_z^{'} = \max \big(  \{|\calI_0^i| \}_{i=1}^{n_o} \big)$ where $\calI_0^i = \{ i \in [n_o] : {\bc^{*}_i}^T \bb_i = 0\}$ and $\bb_i$ is the $i^{\text{th}}$ column of $\bB$. The value $ | \calI_0^{i} |$ is the number of outliers which are orthogonal to $\bc_i^{*}$.  %Similarly, we define vectors $\{ \bc_i^{*} \}_{i=n_o + 1}^{M_2}$ where $\bc_i^{*} = \underset{  \bc^T \ba_{i - n_o} = 1 \: , \: \bc \in \calU^{\perp}}{\arg\min} \: \:  \| \bc^T \bB \|_1$.
\end{definition}

\begin{remark}
In Assumption~\ref{assum_DistUni}, the outliers are randomly distributed. Thus, if $n_o$ is significantly larger than $M_1$, $n_z^{'}$ is significantly smaller than $n_o$ with overwhelming probability.
\end{remark}

%\begin{remark}
%The parameter $\chi$ represents the $\ell_2$-norm of the optimal directions corresponding to the outliers. The value of $\|\bc^{*} \|_2$ can not be large because if $\|\bc^{*} \|_2$ increases, the  value of $\| \bD^{T} \bc^{*}\|_1$ increases.
%In the proposed approach, we can normalize the $\ell_2$-norm of the optimal directions (seating $\bc^{*}$ equal to $\bc^{*}/\| \bc^{*} \|_2$ which means $\chi = 1$) before computing the Innovation Values. Our extensive experiments show that this normalization does not change or improves the performance.
%\end{remark}

\begin{theorem}\label{theo:suffic_random2}
Suppose $\bD$ follows Assumption~\ref{assum_DistUni}. Define $\calA =  \sqrt{\frac{1}{2\pi}} \frac{n_i}{\sqrt{r}} - \sqrt{n_i} - \sqrt{\frac{ n_i \log \frac{1}{\delta}}{2r -2 }}$. If
\begin{eqnarray}
\begin{aligned}
& \calA > 2 n_z^{'} \sqrt{\frac{c_\delta r}{M_1}} + 2 \sqrt{ \frac{n_o  c_\delta r \log n_o/\delta}{M_1}} \: , \\
& \calA > \Bigg[ \frac{n_o }{{M_1}} + \sqrt{\frac{4 n_o }{M_1}} + \sqrt{\frac{2 n_o \log 1/\delta}{(M_1 - 1)M_1}} + \sqrt{\frac{n_o c_{\delta}^{''} \log n_o/\delta }{M_1^2 }}  \\
& + n_z^{'} \sqrt{\frac{c_{\delta}^{''}}{M_1^2}} +
 \sqrt{  \frac{n_o \log n_o/\delta}{M_1^2}+ \frac{\eta_{\delta} \log n_o/\delta}{M_1}   } \Bigg] \sqrt{\frac{ 4 M_1 c_{\delta}}{M_1 - c_{\delta} r}} \:,\\
& \calA > \chi
 \frac{n_o}{\sqrt{M_1}} +
 2\sqrt{n_o} (1 + \sqrt{\chi }) +2 \sqrt{\frac{2 \chi  n_o \log \frac{1}{\delta}}{M -1 }} \:,
\end{aligned}
\label{suf:avalirandom23}
\end{eqnarray}
then (\ref{cond:Innovation_condition})  holds
and  $\calU$ is recovered exactly with probability at least $1 - 7\delta$  where
\begin{align*}
\sqrt{c_{\delta}} =& 3 \max \left(1 ,  \sqrt{\frac{8  M_1 \pi }{(M_1 - 1)r}}
\sqrt{\frac{8 M_1 \log n_0/\delta }{(M_1 - 1)r}} \right),\\
\sqrt{c_{\delta}^{''}} =& 3 \max \left(1 ,  \sqrt{\frac{8  M_1 \pi }{M_1 - 1}} , \sqrt{\frac{16 M_1 \log n_0/\delta }{M_1 - 1}} \right),\\
\eta_{\delta} =& \max \left( \frac{4}{3} \log \frac{2M_1}{\delta} , \sqrt{4 \frac{n_o}{M_1} \log \frac{2 M_1}{\delta}} \right) \:.
\end{align*}
\end{theorem}

\noindent
\textbf{Proof Sketch:} In order to prove  Theorem 1, we leveraged this key feature of the direction of innovation: when $\bd$ is an outlier, $\bc^{*}$ is highly incoherent with $\calU$. Specifically, Theorem ~\ref{theo:suffic_random2} is proved in two steps. First, we derived the sufficient conditions to guarantee that all the optimal directions correspond to the outliers,  $\{ \bc_j^{*} \}_{j=1}^{n_o}$, lie in $\calU^{\perp}$. The exact orthogonality is  not a necessarily condition but it simplifies the analysis.
In fact, the first two sufficient conditions in (\ref{suf:avalirandom23}) guarantee it with high probability. Second, the fact that $\{ \bc_j^{*} \}_{j=1}^{n_o}$ are orthogonal to $\calU$ was leveraged to derive the final conditions to guarantee that (\ref{cond:Innovation_condition})  holds with high probability.

Theorem~\ref{theo:suffic_random2} indicates that when $n_i/r$ is sufficiently larger than $n_o/M_1$, \textcolor{black}{ the proposed approach is guaranteed to detect the randomly distributed outliers exactly. }
It is important to note that in (\ref{suf:avalirandom23}), $n_i$ is scaled with $1/r$ but $n_o$ is scaled with $1/M_1$. It means that if $r$ is sufficiently smaller than $M_1$, iSearch  provably detects the unstructured outliers even if $n_o$ is much larger than $n_i$. The numerical experiments presented in Section~\ref{sec:Num} confirms this feature of iSearch. % and they show that if the outliers are unstructured, iSearch can yield exact recovery even if $n_o > 100\:n_i$. It is important to note that when the outliers are structured, by the definition of outlier, $n_o$ cannot be larger than $n_i$.

\subsection{Structured Outliers}
This section provides the analysis of iSearch with structured outliers. In contrast to the unstructured outliers, structured outliers can form a low dimensional structure different from the structure of the majority of the data points.
Structured outliers are
associated with important rare events  such as malignant tissues~\cite{karrila2011comparison} or web attacks~\cite{kruegel2003anomaly}.
In this section, we assume that the outliers form a cluster outside of $\calU$. The following assumption specifies the presumed model for the distribution of the structured outliers.

\begin{assumption}
A column of $\bB$ is formed as $\bb_i = \frac{1}{\sqrt{1 + \eta^2}} ( \bq + \eta \bv_i)$. The unit $\ell_2$-norm vector $\bq$ does not lie in $\calU$, $\{\bv_i \}_{i=1}^{n_o}$ are drawn uniformly at
random from $\mathbb{S}^{M_1 - 1}$, and $\eta$ is a positive number.
\label{asm:clus}
\end{assumption}

\noindent
According to Assumption~\ref{asm:clus}, the outliers cluster around vector $\bq$ where $\bq \not \in \calU$. In Algorithm 1, if the dimensionality reduction step is performed, the direction search optimization problem is applied to $\bQ^T \bD$. Thus, (\ref{opt:asli1}) is equivalent to
\begin{eqnarray}
\underset{ \bc}{\min} \: \:  \| \bc^T \bD \|_1 \quad \text{subject to} \quad \bc^T \bd = 1 \quad \text{and} \quad \bc \in \calQ \:,
\label{opt:withspan}
\end{eqnarray}
where $\bc \in \mathbb{R}^{M_1 \times 1}$ and $\bD \in \mathbb{R}^{M_1 \times M_2}$. The subspace $\calQ$ is the column-space of $\bD$. In this section, we are interested in studying the performance of iSearch in identifying \textcolor{black}{tightly clustered} outliers because
some of the existing outlier detection algorithms fail  if the outliers form a tight cluster. For instance, the
thresholding based method ~\cite{heckel2013robust} and the sparse representation based algorithm~\cite{soltanolkotabi2012geometric} fail when
 the outliers are close to each other.
Therefore, we assume that the span of $\bQ$ is approximately equal to the column-space of $[\bU \: \: \bq]$. The following Theorem shows that even if the outliers are close to each other, iSearch successfully identifies the outliers provided that  $n_i / \sqrt{r}$ is sufficiently larger than $n_o$.

\begin{theorem}
Suppose the distribution of the inliers/outliers follows Assumption-\ref{assum_DistUni}/Assumption-\ref{asm:clus}. Assume that $\calQ$ is equal to the column-space of $[\bU \:\: \bq]$. Define $\bq^{\perp} = \frac{(\bI -\bU \bU^T)\bq}{ \| (\bI -\bU \bU^T)\bq \|_2}$, define $\beta = \max \left( \{ 1/|\bd_i^T \bq^{\perp}| \: \: : \bd_i \in \bB \} \right)$, define $\bc_i^{*}$ as the optimal point of (\ref{opt:withspan}) with $\bd = \bd_i$, and assume that $\eta < |\bq^T \bq^{\perp}|$.  In addition, Define $\calA = \frac{\sqrt{1 + \eta^2}}{2 \beta} \left( \sqrt{\frac{2}{\pi}} \frac{n_i}{{\sqrt{r}}} - 2\sqrt{n_i} - \sqrt{\frac{2 n_i \log \frac{1}{\delta}}{r -1 }} \right) $.
If
\begin{eqnarray}
\begin{aligned}
& \calA>   n_o \|\bU^T \bq \|_2 + \eta  \sqrt{\frac{n_o r c_{\delta} \log n_o/\delta}{M_1} } \:, \\
& \calA>   n_o |\bq^T \bq^{\perp} | + n_o \eta  \sqrt{\frac{ c_{\delta}'' \log n_o/\delta}{M_1} } \:,
\end{aligned}
\label{eq:theo_2_2c}
\end{eqnarray}
then (\ref{cond:Innovation_condition})  holds and  $\calU$ is recovered exactly with probability at least $1 - 5\delta$.
\label{theor:structured}
\end{theorem}

\noindent
\textbf{Proof Sketch:} Similar to the proof of Theorem~\ref{theo:suffic_random2}, first we derive the sufficient conditions to guarantee that the optimal directions corresponding to the outliers are orthogonal to $\calU$ (the exact orthogonality is  not a necessarily condition but it simplifies the analysis). Subsequently, we leverage the structure of the optimal directions to establish an upper-bound/lower-bound for $\| \bD^T \bc^{*} \|_1$  when $\bd$ is an outliers/inlier.

In sharp contrast to (\ref{suf:avalirandom23}), in (\ref{eq:theo_2_2c}) $n_o$ is not scaled with $1/\sqrt{M_1}$.
Theorem~\ref{theor:structured} indicates that in contrast to the unstructured outliers, the number of the structured outliers should be  sufficiently smaller than the number of the inliers for the small values of $\eta$. This is consistent with our intuition
regarding the detection of structured outliers. \textcolor{black}{ If the columns of $\bB$ are highly structured  and most of the data points are outliers, it violates the definition of outlier to label the columns of $\bB$ as outliers.} \\
\textcolor{black}{The presence of parameter $\beta$ emphasizes that the closer the outliers are to $\calU$, the harder it is to distinguish them.} %The challenging step when  the outliers are very close to $\calU$ is finding a proper direction of Innovation. We desire the direction of Innovation to lie in or be close to $\calU^{\perp}$. If the outlier is very close $\calU$ it is hard for a direction in $\calU^{\perp}$ to satisfy the linear constraint of (\ref{opt:asli1}).
In Section~\ref{sec:Num}, it is shown that iSearch significantly
outperforms the existing methods when
 the outliers  are  close to $\calU$. The main reason is that even if an outlier is close to $\calU$, its corresponding optimal direction obtained by (\ref{opt:asli1})  is highly incoherent with $\bU$. % Therefore, its corresponding optimal direction is incoherent with the inliers.

%For instance, in the clustering error detection experiment (presented in Section~\ref{sec:EXP_real_motin}) in which the structured outliers are very close to $\calU$, iSearch significantly outperforms the existing methods including CoP. The main reason is that when an outlying column is close to $\calU$, it will have a high Coherence Value because it has strong inner-product values with many of the inliers. In the same setting, iSearch finds a direction which is highly incoherent with $\bU$. Therefore, even if the outlier is close to $\calU$, its corresponding direction of innovation is highly incoherent with $\calU$.

\subsection{Linearly Dependent Outliers}

In some applications, the outliers are linearly dependent. For instance, in~\cite{gitlin2018improving}, it was shown that a robust PCA algorithm can be used to reduce the clustering error of a subspace segmentation method.
In this application, a small subset of the outliers can be linearly dependent.
The following assumption specifies the presumed model for matrix $\bB$.

\begin{assumption}
Define
subspace $\calU_o$ with dimension $r_o$ such that $\calU_o \notin \calU$ and $\calU \notin \calU_o$.
The outliers are randomly distributed on  $\mathbb{S}^{M_1 - 1} \cap \calU_o$. The orthonormal matrix $\bU_o \in \mathbb{R}^{M_1 \times r_o}$ is a basis for $\calU_o$.
\label{asm:out}
\end{assumption}

\noindent
The following theorem establishes the sufficient conditions to guarantee the performance of iSearch with linearly dependent outliers. The procedure to proof the following theorem is similar to the previous theorems.  The main difference was the techniques used to bound the value of $\|\bD^T \bc^{*} \|_1$.

\begin{theorem}
Suppose the distribution of the inliers/outliers follows Assumption-\ref{assum_DistUni}/Assumption-\ref{asm:out}. Define $\calA = \sqrt{\frac{2}{\pi}} \frac{n_i}{\sqrt{r}} - 2\sqrt{n_i} - \sqrt{\frac{2 n_i \log \frac{1}{\delta}}{r -1 }}$. If
\begin{eqnarray}
\begin{aligned}
& \calA > 2 n_z^{'} \| \bU^T \bU_o \| + 2 \| \bU^T \bU_o \| \sqrt{n_o  \log n_o/\delta } \: , \\
& \calA > \frac{2\| {\bU}^T \bU_o \|}{\xi} \Bigg( \frac{n_o }{\sqrt{r_o}} +
2\sqrt{n_o } + \\
&\hspace{0.3in}\sqrt{\frac{2 n_o \log \frac{1}{\delta}}{r_o -1 }} + 2\sqrt{  \left( \frac{n_o}{r_o} + \eta_{\delta}^{'} \right)\log \frac{n_o}{\delta}  }  +n_z^{'} \Bigg) \:,\\
& \calA > \left(\frac{ \chi n_o}{\sqrt{r_o}} + 2\sqrt{ \chi n_o} +\sqrt{ \chi \frac{2 n_o \log \frac{1}{\delta}}{r_o -1 }} \right) \| \bU_o^T \bU^{\perp} \| \: ,
\end{aligned}
\label{eq:hjg3}
\end{eqnarray}
then (\ref{cond:Innovation_condition})  holds and $\calU$ is recovered exactly with probability at least $1 - 5 \delta$ where $\eta_{\delta}^{'} = \max \left(\frac{4}{3} \log 2 (r_o) /\delta \: , \: \sqrt{4 \frac{n_o}{r_o} \log \frac{2 r_d}{\delta}} \right) $ and $\xi = \frac{\min \left( \left\{ \|\bb_j^T \bU^{\perp} \|_2 \right\}_{j=1}^{n_o} \right)}{\|\bU_o^T \bU^{\perp} \|}$.
\label{theo:linear_dep}
\end{theorem}

\noindent
 \textcolor{black}{Theorem~\ref{theo:linear_dep} indicates that   $n_i/r$ should be sufficiently larger} than \textcolor{black}{ $n_o/r_o$}. If $r_o$ is comparable to $r$, it is in fact a necessary condition because we can not label the columns of $\bB$ as outliers if $n_o$ is also comparable with $n_i$. If $r_o$ is large,  the sufficient condition is similar to the sufficient conditions of Theorem~\ref{theo:suffic_random2} in which the outliers are distributed randomly on $\mathbb{S}^{M_1 -1 }$.
 %\textcolor{black}{In the proof of Theorem~\ref{theo:linear_dep}, it is shown that if $\bd$ is an outlier and  the first two sufficient condition of (\ref{eq:hjg3}) are satisfied, the optimal point of (\ref{opt:asli1}) lies in $\calU^{\perp}$ with high probability. The factor $ \| \bU^T \bU_o \|$ shows that the closer $\calU_o$ is  to $\calU^{\perp}$, it is more likely that the optimal point of (\ref{opt:asli1}) lies in $\calU^{\perp}$. The reason is that if $\bd$ is closer to $\calU^{\perp}$, it is easier for a direction in $\calU^{\perp}$ to satisfy the linear constrain $\bc^T \bd = 1$.}
It is also informative to compare the requirements of iSearch with the requirements of CoP.
With iSearch,  $n_i/{r}$ should be sufficiently larger than $ \frac{n_o}{r_o} \| \bU_o \bU^{\perp} \|$ to guarantee that the algorithm distinguishes the outliers successfully. With CoP, $n_i/r_i$ should be sufficiently larger than $n_o/r_o + \| \bU_o^T \bU \| n_i/r_i$~\cite{rahmani2017coherence22f,gitlin2018improving}. The reason that CoP requires a stronger condition is that iSearch finds a direction for each outlier which is highly incoherent with $\calU$.

\subsection{Outlier Detection When the Inliers are Clustered}
In the analysis of the robust PCA methods, mostly it is assumed that the inliers are randomly distributed in $\calU$. In practise the inliers form several clusters in $\calU$. In this section, it is assumed that the inliers form $m$ clusters.
The following assumption specifies the presumed model and Theorems~\ref{theo:suffic_random24} provides the sufficient conditions.

\begin{assumption}
The matrix of inliers can be written as $\bA = [\bA_1 \: ... \: \bA_m] \bT_A$ where $\bA_k \in \mathbb{R}^{M_1 \times {n_i}_k}$, $\sum_{k=1}^{m}  {n_i}_k = n_i$, and $\bT_A$ is an arbitrary permutation matrix.
The columns of $\bA_k$  are drawn uniformly at random from the
intersection of  subspace $\calU_k$ and $\mathbb{S}^{M_1-1}$ where $\calU_k$ is a $d$-dimensional subspace. In other word, the columns of $\bA$ lie in a union of subspaces $\{ \calU_k \}_{k=1}^m$ and $\left(\calU_1 \oplus \: ... \oplus \calU_m \right)= \calU$ where $\oplus$ denotes the direct sum operator.
\label{asm:union_of_sunb}
\end{assumption}

\begin{theorem}
Suppose the distribution of the outliers/inliers follows Assumption-\ref{assum_DistUni}/Assumption-\ref{asm:union_of_sunb}. Define
\begin{eqnarray*}
\begin{aligned}
& g = \argmin_{k}    \underset{\delta \in \calU_k  \atop \| \delta \| = 1}{\inf}  \| \delta^T \bA_k \|_1 \:, \\
& \rho = \underset{\delta \in \calU  \atop \| \delta \| = 1}{\inf} \sum_{k=1}^{m} \| \delta^T \calU_k \|_2 \:, \\
& \calA =  \rho \left( \sqrt{\frac{2}{\pi}} \frac{n_g}{\sqrt{d}} - 2\sqrt{n_g} - \sqrt{\frac{2 n_g \log \frac{1}{\delta}}{r -1 }}  \right) \:. \\
\end{aligned}
\end{eqnarray*}
If the sufficient conditions in (\ref{suf:avalirandom23}) are satisfied, then the inequality (\ref{cond:Innovation_condition})  holds and  $\calU$ is recovered exactly with probability at least $1 - 7\delta$.
\label{theo:suffic_random24}
\end{theorem}

\noindent
\textbf{Proof sketch:} The difference between Theorem~\ref{theo:suffic_random24} and Theorem~\ref{theo:suffic_random2} is in their presumed model for the distribution of the inliers. The procedure used to prove Theorem~\ref{theo:suffic_random24} is similar to the procedure used in the proof of Theorem~\ref{theo:suffic_random2} but it required new techniques to bound $\| \bA^T \bc^{*} \|$.

\smallbreak
In Assumption~\ref{asm:union_of_sunb}, the dimensions of the subspaces $\{\calU_k \}_{k=1}^{m}$ are equal and the distribution of the inliers inside these subspace are similar. Therefore, we can roughly say
$g = \argmin_k {n_i}_k$~\cite{lerman2015robustnn}. Thus, the sufficient conditions indicate that the population of the smallest cluster scaled by $1/\sqrt{d}$
should be sufficiently larger than $n_o/M_1$. The parameter $\rho=\underset{\delta \in \calU  \atop \| \delta \| = 1}{\inf} \sum_{k=1}^{m} \| \delta^T \calU_k \|_2 \:$ is similar to the permeance statistic introduced in~\cite{lerman2015robustnn}. It shows  how well the inliers are distributed in $\calU$. Evidently, if the inliers populate all the directions inside $\calU$,
 a subspace recovery algorithm is more likely to recover $\calU$ correctly.
\textcolor{black}{
However, having a large value of  permeance statistic is not a necessary condition}. The reason that permeance statistic appears in the sufficient conditions is that we establish the sufficient conditions to guarantee the performance of iSearch in the worst case scenarios.
\textcolor{black}{ In fact, if the inliers are close to each other or the subspaces $\{\calU_i \}_{i=1}^{m}$ are close to each other, generally the performance of iSearch improves because the more  inliers are close to each other, the smaller  their Innovation Values are. }

\subsection{Noisy Data}
\label{sec:nooise}
The key feature of the proposed approach is that when $\bd$ is an outlier, $\bc^{*}$ is highly incoherent with $\calU$. In the proof of the presented theorems, we derived sufficient conditions to guarantee that $\bc^{*}$ is orthogonal to $\calU$. It is not a necessarily condition but it simplified the analysis.
When the inliers are noisy, we can not find a direction which is orthogonal to all of them because they do not lie in a low dimensional subspace. In this section, first we focus on showing that even if the data is noisy, the direction of innovation corresponding to an outlier remains incoherent with $\calU$.  In the following theoretical results, it is assumed that the data follows Assumption~\ref{asm:noiss}. According to Assumption~\ref{asm:noiss}, each inlier is added with a random direction and  each data
column has an expected squared norm equal to 1.

\begin{assumption}
The matrix $\bD$ can be expressed as
$
\bD =  [\bB \hspace{.2cm} \frac{1}{1+\sigma_n^2}(\bA+\bE)] \: \bT \:.
$
The matrix $\bE \in \mathbb{R}^{M_1 \times n_i}$ represents the presence of noise and it can be written as $\bE = \sigma_n \bN$ where the columns of $\bN \in \mathbb{R}^{M_1 \times n_i}$  are drawn  uniformly at random from $\mathbb{S}^{M_1 - 1}$ and $\sigma_n$ is a positive number which controls the power of the added noise.
\label{asm:noiss}
\end{assumption}

\begin{lemma}
Suppose  $\bD$ follows Assumption~\ref{asm:noiss} and assume that $\bA$ and $\bB$ follow Assumption~\ref{assum_DistUni}.  Define $\varrho = \max_i \{ \| \bU^T \bc_i^{*} \|_2 \}_{i=1}^{n_o}$, $\upsilon = \max \{ \frac{1}{ \| \bd^T \bR \|_2} \}$ , and
\begin{eqnarray*}
\begin{aligned}
& \calA_e = \frac{1 + \upsilon}{1+ \sigma_n^2} \left(\frac{  n_i}{ \sqrt{M_1 -0.5} }  + 2\sqrt{ n_i} + \sqrt{\frac{2 n_i \log \frac{1}{\delta}}{M_1 -1 }} \right)\:, \\
& \calA_i = \frac{1}{1+ \sigma_n^2} \left(\frac{ \: n_i}{ \sqrt{r} } \sqrt{\frac{2}{\pi}} - 2\sqrt{ \: n_i} - \sqrt{\frac{2  \: n_i \log \frac{1}{\delta}}{r -1 }} \right) \:, \\
& \calD_o = (\upsilon-1) \frac{n_o}{\sqrt{M_1 - 0.5} } \sqrt{\frac{2}{\pi}} +  2(1+ \upsilon)\sqrt{n_o} +  \\
&\hspace{0.4in}(1+ \upsilon)\sqrt{\frac{2 n_o \log \frac{1}{\delta}}{M_1 -1 }} \:,
 \end{aligned}
\end{eqnarray*}
where $\bR$ is an orthonormal basis for $\calU^{\perp}$. Then
\begin{eqnarray}
\varrho \le \frac{\sigma_n \calA_e + \calD_o}{\calA_i}
\label{eq:bounded_rho}
\end{eqnarray}
with probability at least $1- 4\delta$.
\label{them:stability}
\end{lemma}

\noindent
Lemma~\ref{them:stability} establishes a upper-bound for $\varrho = \max_i \{ \| \bU^T \bc_i^{*} \|_2 \}_{i=1}^{n_o}$. In practise, $\varrho$ is much smaller than the value of the upper-bound because in the proof we consider  the  worst  case  scenarios.  Lemma~\ref{them:stability} indicates that the direction of innovation corresponding to an outlier stays incoherent with $\calU$ provided that $n_i (M_1/r)$ is sufficiently large. The following theorem provides the sufficient conditions to guarantee the performance of iSearch with noisy data.

\begin{theorem}
Suppose  $\bD$ follows Assumption~\ref{asm:noiss} and assume that $\bA$ and $\bB$ follow Assumption~\ref{assum_DistUni}.  Define $\calA =  \sqrt{\frac{1}{2\pi}} \frac{n_i}{\sqrt{r}} - \sqrt{n_i} - \sqrt{\frac{ n_i \log \frac{1}{\delta}}{2r -2 }}$ and $\upsilon = \max \{ \frac{1}{ \| \bd^T \bR \|_2} \}$. If \begin{eqnarray*}
\begin{aligned}
& \frac{1}{1 + \sigma_n^2} \calA \ge (\upsilon - \frac{1}{1+ \sigma_n}) \frac{n_o}{\sqrt{M_1 - 0.5} } \sqrt{\frac{2}{\pi}} + 4\upsilon\sqrt{n_o} + \\
&2\upsilon \sqrt{\frac{2 n_o \log \frac{1}{\delta}}{M_1 -1 }} + \sigma_n \left( \chi^{'} + \frac{\upsilon}{1+ \sigma_n^2}\right)\Bigg( \frac{n_i}{\sqrt{M_1 - 0.5} } \sqrt{\frac{2}{\pi}}  +\\
&\hspace{5cm} 2\sqrt{n_i} + \sqrt{\frac{2 n_i \log \frac{1}{\delta}}{M_1 -1 }} \Bigg)
\end{aligned}
\end{eqnarray*}
then (\ref{cond:Innovation_condition}) holds with probability at least $1- 5\delta$.
\label{thm:final}
\end{theorem}

\noindent
In Theorem~\ref{thm:final}, the outliers are unstructured and its sufficient conditions are similar to the sufficient conditions of Theorem~\ref{theo:suffic_random2}. Roughly, Theorem~\ref{thm:final} states that if $\frac{n_i}{(1+\sigma_n^2)\sqrt{r}}$ is sufficiently larger than $\frac{n_o}{\sqrt{M_1}}$, Innovation Values can reliably be used to distinguish the outliers. For clustered outliers and  linearly dependent outliers, similar guarantees can be established.

\section{Numerical Experiments}\label{sec:Num}
In this section,
a set of experiments with synthetic data and real data are presented to study
the performance and the properties of the iSearch algorithm.
In the presented experiments,  iSearch is compared with the existing methods including FMS~\cite{lerman2014fast}, GMS~\cite{zhang2014novel}, CoP~\cite{rahmani2017coherence22f}, OP~\cite{lamport10}, and R1-PCA~\cite{lamport21}.

\subsection{Phase Transition}
\label{sec:phase_t}
\begin{figure}[h!]
\begin{center}
\mbox{\hspace{-0.15in}
\includegraphics[width=1.95in]{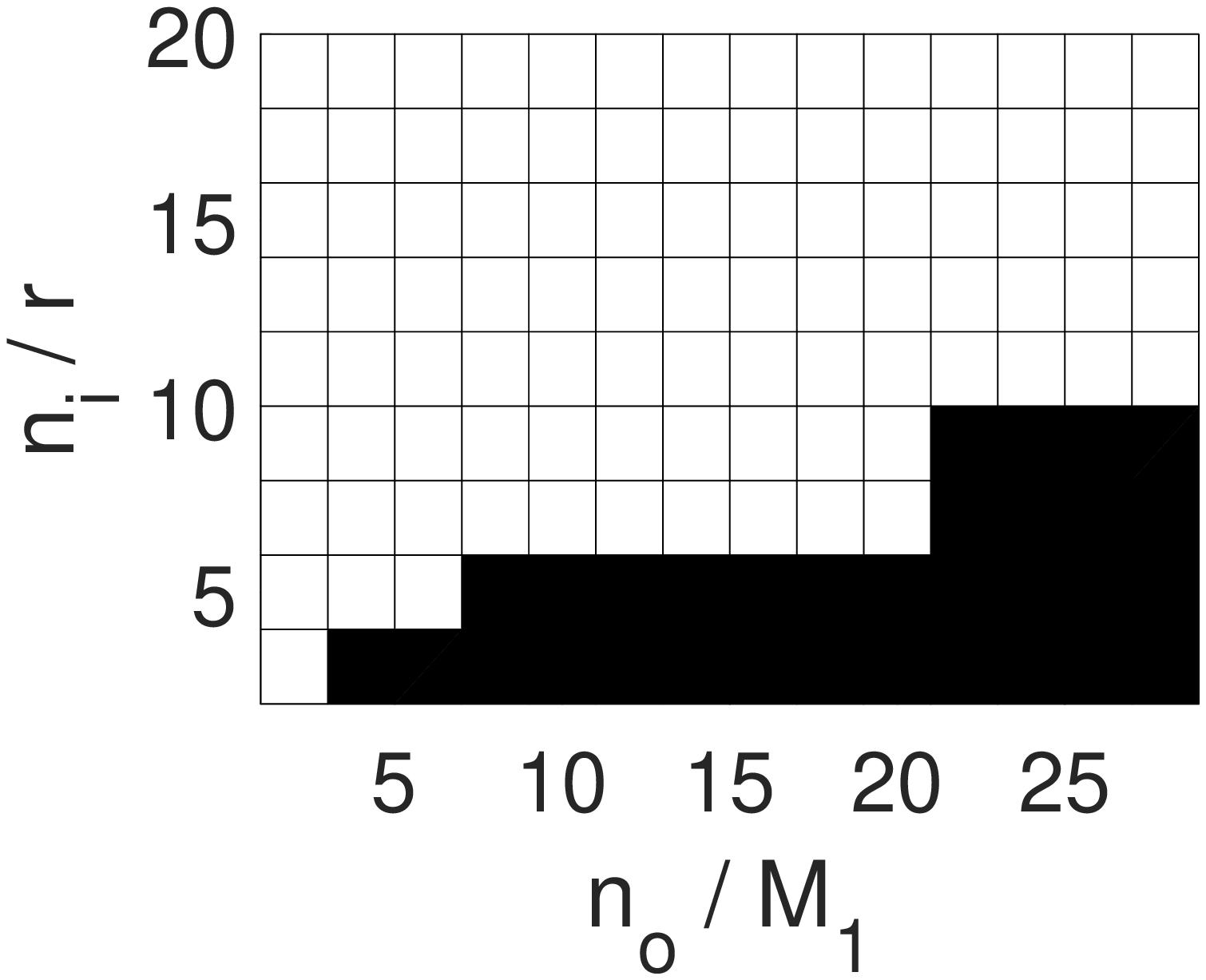}\hspace{-0.15in}
\includegraphics[width=1.95in]{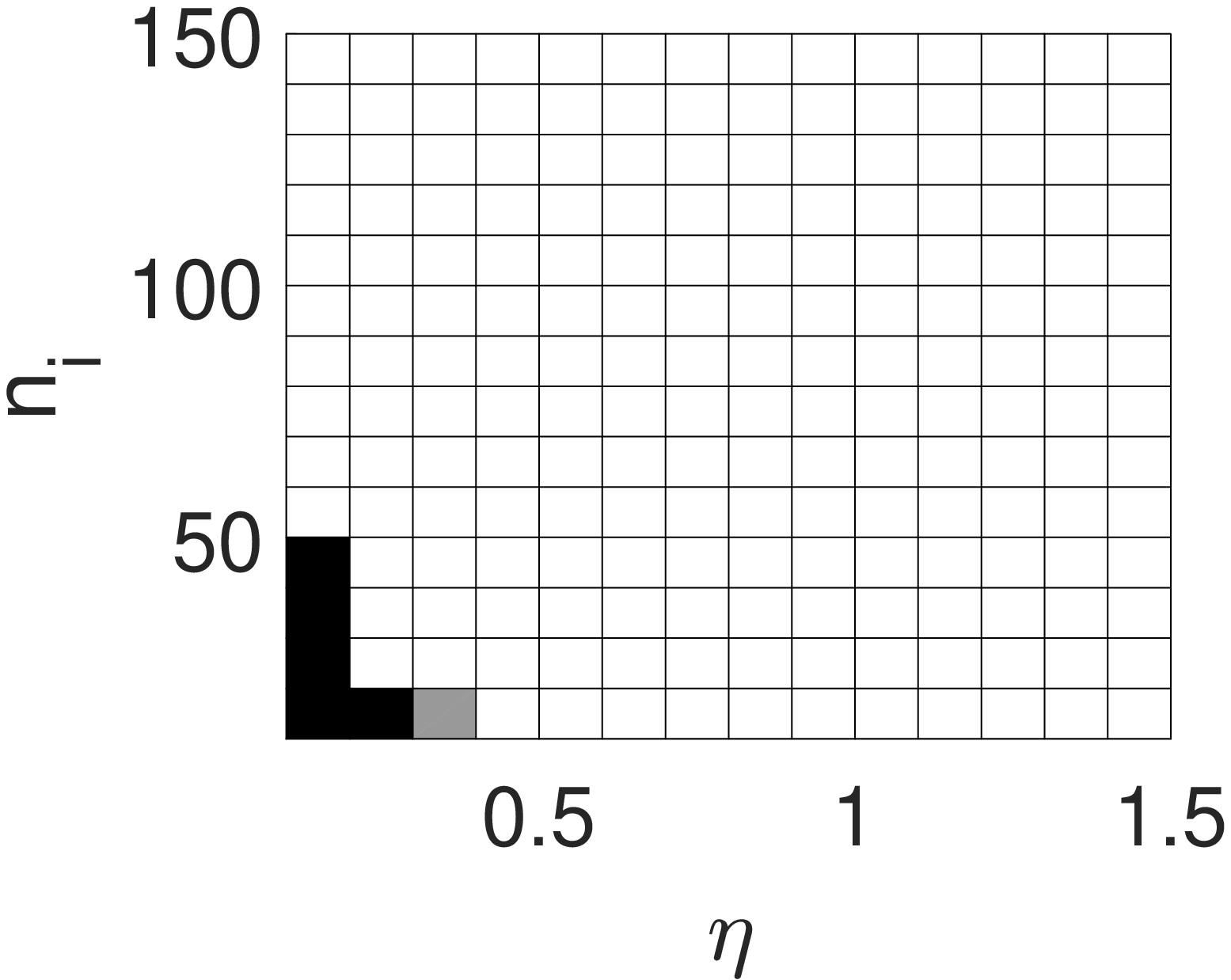}
}
\end{center}
\vspace{-0.15in}
           \caption{ Left plot: the  phase transition of iSearch in presence of the unstructured outliers versus $n_i/r$ and $n_o/M_1$. White indicates correct
subspace recovery and black designates incorrect recovery. Right plot: phase transition with structured outliers versus $n_i$ and $\eta$. In both plots, $M_1 = 100$ and $r = 4$. }
    \label{fig:phase_tr}

\end{figure}

In this experiment,
the  phase transition of the proposed approach is studied. First, it is shown that if $n_i/r$ is sufficiently large,
\textcolor{black}{
iSearch can successfully recover the span of the inliers even in the dominant presence of the unstructured outliers.}
Define $\hat{\bU}$ as an orthonormal basis for the recovered subspace. A trial is considered successful if
\begin{eqnarray*}
\frac{\| (\bI - \bU \bU^T) \hat{\bU} \|_F}{ \|\bU \|_F } < 10^{-2} \: .
\end{eqnarray*}
The data follows Assumption~\ref{assum_DistUni} with $r = 4$ and $M_1 = 100$. Fig. ~\ref{fig:phase_tr} shows the phase transition of iSearch versus $n_i/r$ and $n_o/M_1$. White indicates correct
subspace recovery and black designates incorrect recovery.
Theorem~\ref{theo:suffic_random2} indicated that if $n_i/r$ is sufficiently large, iSearch yields exact recovery even if $n_o$ is  larger than $n_i$. This experiment confirms the theoretical result.  \textcolor{black}{ According to Fig.~\ref{fig:phase_tr}, even when $n_o = 3000$, 40 inliers are enough to guarantee exact subspace recovery. } In the second plot of Fig.~\ref{fig:phase_tr}, the outliers are structured and the distribution of outliers follows Assumption~\ref{asm:clus} with $n_o=25$.  One can observe that even when $\eta$ is small, a sufficiently large value of $n_i$ guarantees the performance of iSearch.

\begin{figure}[h!]
\begin{center}
\mbox{
\includegraphics[width=2.7in]{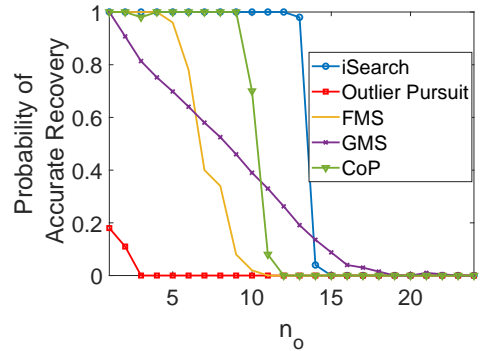}
}
\end{center}
\vspace{-0.15in}
           \caption{ \textcolor{black}{The probability of accurate subspace recovery versus the number of structured outliers ($n_i=100$, $\eta=0.1$, $M_1 = 100$, and $r=10$). } A trial is considered successful if
$
\frac{\| (\bI - \bU \bU^T) \hat{\bU} \|_F}{ \|\bU \|_F } < 10^{-2}
$ where $\hat{\bU}$ is an orthonormal basis for the recovered subspace. }
\label{fig:new_structured}

\end{figure}

\subsection{Structured Outliers}\label{sec:EXP_struc_syn}

In this experiment, we consider structured outliers. The distribution of the outliers follows Assumption~\ref{asm:clus} with $\eta = 0.1$ and $M_1 = 100$.
In addition, the inliers are clustered and they lie in a union of 5 2-dimensional linear subspaces. There are 20 data points in each subspace (i.e., $n_i=100$) and $r=10$. A successful trial is defined similar to Section~\ref{sec:phase_t}. We are interested in investigating the performance of iSearch in identifying structured outliers when they are close to $\calU$. Therefore, we generate vector $\bq$, the center of the cluster of the outliers, close to $\calU$. Vector $\bq$ is constructed as
\begin{eqnarray*}
\bq = \frac{\left[\bU \: \: \bp \right]\bh}{ \left\|\left[\bU \: \: \bp \right] \bh  \right\|_2} \:,
\end{eqnarray*}
where the unit $\ell_2$-norm vector $\bp \in \mathbb{R}^{M_1 \times 1}$ is generated as a random direction on $\mathbb{S}^{M_1 -1}$ and the elements of $\bh \in \mathbb{R}^{(r+1) \times 1}$ are sampled independently from $\calN(0,1)$. The generated vector $\bq$ is close to $\calU$ with high probability because the column-space of $\left[\bU \: \: \bp \right]$ is close to the column-space of $\bU$.  Fig.~\ref{fig:new_structured} shows the probability of accurate subspace recovery versus the number of outliers. The number of evaluation runs was 50. One can observe that in contrast to the unstructured outliers, the robust PCA methods tolerate few number of structured outliers and iSearch exhibits higher robustness to the presence of the structured outliers.

\subsection{Noisy Data}

\begin{figure}[h!]
\begin{center}
\mbox{
\includegraphics[width=2.7in]{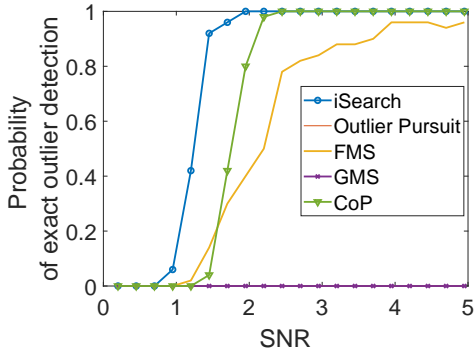}
}
\end{center}
\vspace{-0.15in}
           \caption{ The probability of exact outlier detection versus SNR. The data contains 10 structured outliers and 300 unstructured outliers ($n_i=100$, $n_o=310$, $r=5$, and $M_1 = 100$). }
\label{fig:SNR_NW}
\end{figure}

In this section, we consider the simultaneous presence of noise,  the structured outliers and the unstructured outliers. In this experiment, $M_1 = 100$, $r = 5$, and $n_i = 100$. The data contains 300 unstructured and 10 structured outliers. The distribution of the structured outliers follow Assumption~\ref{asm:clus} with $\eta = 0.1$. The vector $\bq$, the center of the cluster of the structured outliers, is generated as a random direction on $\mathbb{S}^{M_1 - 1}$. The generated data in this experiment can be expressed as $\bD = [\bB \: \: (\bA+\bE)]$ (the matrix $\bE$ represents the additive Gaussian noise).
Since the data is noisy, the algorithms can not achieve exact subspace recovery. Therefore,  we examine the probability that an algorithm distinguishes  all the outliers correctly. Define vector $\mathbf{f} \in \mathbb{R}^{M_2 \times 1}$ such that $\mathbf{f}(k) = \| (\bI - \hat{\bU} \hat{\bU}^T) \bd_k \|_2$. A trial is considered successful if
\begin{eqnarray*}
\max \bigg( \{\mathbf{f}(k) \: \: : \: \: k > n_o \} \bigg) < \min \bigg( \{\mathbf{f}(k) \: \: : \: \: k \le n_o \} \bigg) \:.
\end{eqnarray*}
Fig.~\ref{fig:SNR_NW} shows the probability of exact outlier detection versus SNR which is defined as $\text{SNR} = \frac{\| \bA \|_F^2}{\| \bE \|_F^2 }$. It shows that iSearch robustly distinguishes the outliers in the strong presence of noise. The number of evaluation runs was 50.

\subsection{Innovation Value vs Coherence Value}
\begin{figure}[h!]
 \centering
    \includegraphics[width=0.48\textwidth]{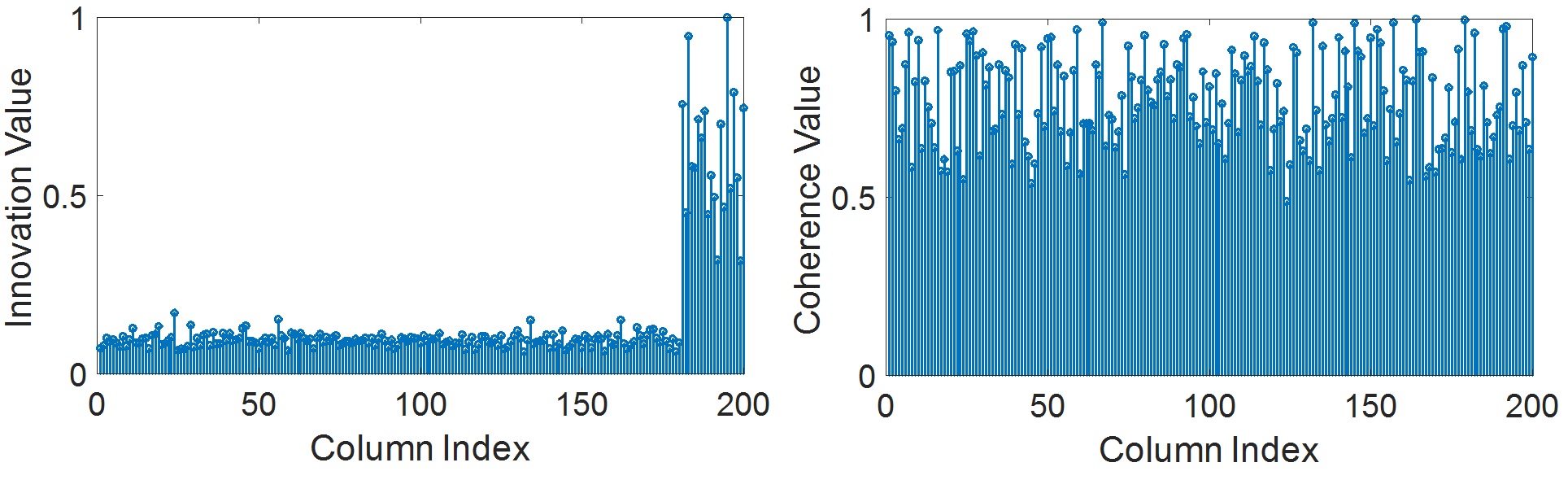}
    \vspace{ - .05 cm}
    \caption{ This figure compares the vector of the Coherence Values (right plot) with the vector of the Innovation Values (left plot). The last 20 columns are the outliers. }
    \label{fig:inno_with_coh}
\end{figure}

We simulate a scenario in which the outliers are randomly distributed but they are close to the span of the inliers.
Suppose $r=8$, $n_i = 180$, and $n_o = 20$. The outliers are generated as $[\bU \:\: \bH]\: \bG$ where $\bH \in \mathbb{R}^{M_1 \times 2}$ spans a random 2-dimensional subspace and the elements of $\bG \in \mathbb{R}^{10 \times 20} $ are sampled independently from $\calN(0,1)$.  Fig.~\ref{fig:inno_with_coh} compares  Innovation Values with Coherence Values. The last 20 columns are the outliers.
One can observe that the Coherence Values can not make the outliers distinguishable from the inliers. The main reason is that in contrast to Coherence Pursuit, iSearch finds a direction corresponding to each outlier which is strongly incoherent with $\calU$.

\subsection{Clustered Inliers}
 In practice, the inliers  are not necessarily distributed uniformly at random and they are mostly close to each other and they form one or multiple clusters. In this experiment, it is shown that in contrast to most of the existing methods, if the inliers form a cluster in $\calU$, the performance of iSearch improves. It is assumed that the distribution of inliers follows Assumption~\ref{asm:inliers_clus}.
 \begin{assumption}
Each column of matrix $\bA$ is formed as $\ba_i = \frac{ \bU \bs_i}{\|  \bU \bs_i \|_2} $ where $\bs_i =  \bw + \gamma \bz_i$. The vector $\bw \in \mathbb{R}^{r \times 1}$ is a unit $\ell_2$-norm vector and $\{\bz_i \}_{i=1}^{n_o}$ are drawn uniformly at
random from $\mathbb{S}^{r - 1}$.
\label{asm:inliers_clus}
\end{assumption}

\begin{figure}[h!]
\begin{center}
\mbox{
\includegraphics[width=2.5in]{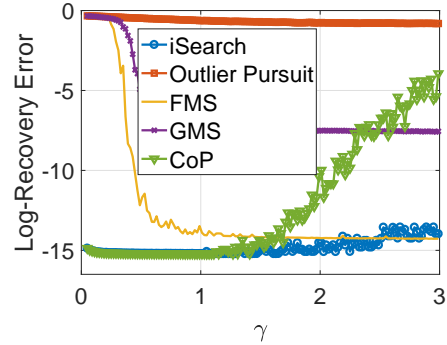}
}
\end{center}`
\vspace{-0.15in}
           \caption{ The Log-Recovery Error of the robust PCA methods versus  $\gamma$. The parameter $\gamma$ is defined in Assumption~\ref{asm:inliers_clus}.
           Log-Recovery Error is defined as
           $
\log_{10} \left( \frac{\| (\bI - \bU \bU^T) \hat{\bU} \|_F}{ \|\bU \|_F } \right).
$
           When $\gamma$ is small, the inliers are close to each other and when it is larger, the distribution of the inliers is similar to the distribution of the inliers in Assumption~\ref{assum_DistUni} (i.e., uniformly at random on $\calU \cap \mathbb{S}^{M_1 -1}$).}
\label{fig:clus_inlierss}
\end{figure}

\begin{figure*}[t!]
	\centering
    \includegraphics[width=0.75\textwidth]{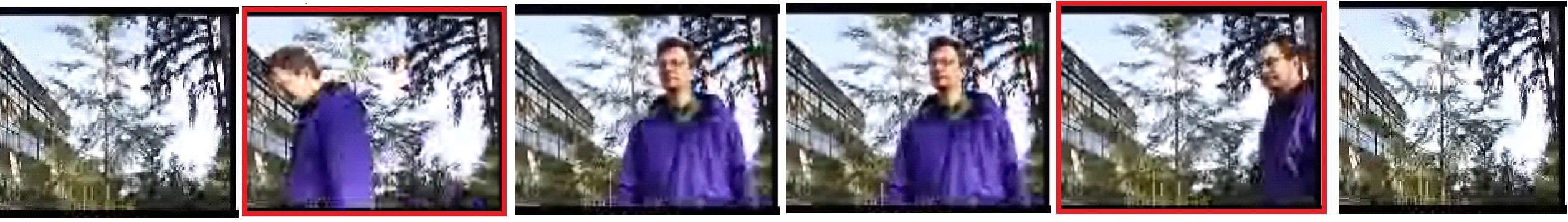}
    \vspace{-.02in}
    \caption{Some of the frames of the Waving Tree video file. The highlighted frames are detected as outliers by R1-PCA.  }
    \label{fig:activity}
\end{figure*}
\noindent
The parameter $\gamma$ controls how close the inliers are to each other. If $\gamma$ goes to infinity, the distribution of the inliers converges to the distribution of the inliers in Assumption~\ref{assum_DistUni}. In this experiment, $n_i=20$, $M_1=100$, and $n_o=20$. The distribution of the outliers follows Assumption~\ref{asm:out} with $r_o=20$ and the dimension of intersection between $\calU_o$ and $\calU$ is equal to 5. This robust PCA algorithms are used to obtain a basis for the dominant subspace (the span of the inliers). Fig.~\ref{fig:clus_inlierss} shows Log-Recovery-Error defined equal to $
\log_{10} \left( \frac{\| (\bI - \bU \bU^T) \hat{\bU} \|_F}{ \|\bU \|_F }  \right)
$
versus parameter $\gamma$.  The number of evaluation runs was 100. One can observe that the performance of most of the robust PCA methods degrade when $\gamma$ is small. The reason is that if the value of $\gamma$ increases, the distribution of the inliers becomes closer to the uniformly at random distribution on $\calU \cap \mathbb{S}^{M_1 -1}$). In contrast,  Coherence Pursuit and iSearch yield better performance  when the inliers form a cluster (the smaller values of $\gamma$). The reason is that if the inliers become closer to each other, their Coherence-Values/Innovation-Values increase/decrease because their similarities increase and they are less innovative.

\subsection{Detecting Outliers in Real Data}
\label{sec:EXP_real_motin}
An application of the outlier detection methods is to identify the misclassified data points of a clustering method~\cite{gitlin2018improving,rahmani2017coherence22f}.
In each identified cluster, the misclassified data points can be considered as outliers. In this experiment, we assume an imaginary clustering method whose clustering error is 25 $\%$. The robust PCA method is applied to each  cluster to find the misclassified data points.
The clustering is re-evaluated after identifying the misclassified data points. Algorithm 2 adapted from~\cite{rahmani2017coherence22f} shows how a robust PCA algorithm is used to detect the misclassified data points and update the identified clusters.
We use the Hopkins155 dataset~\cite{tron2007benchmark}, which
contains data matrices with 2 or 3 clusters. In this experiment, 27 matrices with 3 clusters are used (i.e., the columns of each data matrix lie in 3 clusters). The outliers are linearly dependent and they are very close to the span of the inliers since the clusters in the Hopkins155 dataset are close to each other. In addition, the inliers form a tight cluster. Evidently, the robust PCA methods  which assume that the outliers are randomly distributed  fail in this task.
This experiment with real data contains most of the challenges that a robust PCA method can encounter. For more details about this experiment, we refer the reader to~\cite{rahmani2017coherence22f,gitlin2018improving}.  Table~\ref{tab:accuracy} shows the average clustering error after applying the robust PCA methods to the output of the clustering method.
One can observe that iSearch significantly outperforms the other methods. The main reason is that iSearch is robust against outliers which are closed to $\calU$. In addition, the coherency between the inliers enhances the performance of iSearch.

\begin{algorithm}
\caption{Subspace Clustering Error Correction Using Robust PCA}
{
\textbf{Input}: The matrices $\{ \hat{\bD}^i \}_{i = 1}^L$ are the clustered data (the output of a subspace clustering algorithm) and $L$ is the number of clusters.
\smallbreak
\textbf{1. Finding a basis for each cluster.}\\
Apply the robust PCA algorithm to the matrices $\{  \hat{\bD}^i \}_{i = 1}^L$. Define the orthonormal matrices $\{  \hat{\bU}^i \}_{i = 1}^L$ as the learned bases for the inliers of $\{  \hat{\bD}^i \}_{i = 1}^L$, respectively.

\textbf{2. Cluster the data with the obtained bases. }
Update the data clustering with respect to the obtained bases $\{ \hat{\bU}^i \}_{i = 1}^L$ (the matrices $\{ \hat{\bD}^i \}_{i = 1}^L$ are updated), i.e., data point $\bd$ is assigned to the $i^{\text{th}}$ cluster if $ i = \underset{k}{ \argmax} \: \| \bx^T \hat{\bU}^k \|_2$.
\smallbreak
\textbf{Output:} The matrices $\{ \hat{\bD}^i \}_{i = 1}^L$ are the identified data clusters.
}
\end{algorithm}

\begin{table}[h]
\centering
\caption{Clustering error after using the robust PCA methods to detect the misclassified data points. }
\begin{tabular}{| c | c | c  | c |c|}
\hline
iSearch & CoP &     FMS  & R1-PCA  & PCA \\
 \hline
  2 $\%$ & 7 $\%$ & 20.3 $\%$ & 16.8 $\%$  &  12.1 $\%$ \\
  \hline
\end{tabular}
\label{tab:accuracy}
\end{table}

\subsection{Activity Detection in Real Noisy Data}
\label{sec:EXP_real_motin}
In this experiment, we use the robust PCA methods to identify a rare event in a video file. We use the Waving Tree video file~\cite{li2004statistical}. In this video,
 a tree is smoothly waving and in
the middle of the video a person crosses the frame. The frames which only contain the background (the tree and the environment) are inliers and the few frames corresponding to the event, the presence of the person, are the outliers. Since the tree is waving, the inliers are noisy and we use $r=3$ for all the methods. In addition, we identify column $\bd$ as outlier if ${\| \bd - \hat{\bU}\hat{\bU} \bd\|_2}/{\| \bd \|_2} \ge 0.2$ where $\hat{\bU}$ is the recovered subspace. In this experiments, the outliers are very similar to each other since the  consecutive frames are quite similar to each other. We use iSearch, CoP, FMS, and R1-PCA to detect
the outlying frames. iSearch, CoP, and FMS identified all the
outlying frames correctly.  R1-PCA could not identify those frames in which the person does not move. The reason is that those frames are exactly similar to each other. Fig.~\ref{fig:activity} shows some of the outlying frames which were missed by R1-PCA.

\section{Conclusion}
A new robust (to outlier) PCA method, termed iSearch, was proposed which uses a convex optimization problem to measure the innovation of the data points. The proposed approach recovers the span of the inliers using the least innovative data points. It was shown that iSearch can provably recover the span of the inliers with different models for the distribution of the outliers. In addition, analytical performance guarantees for iSearch with clustered inliers were presented. It was shown that finding the optimal directions  makes iSearch significantly robust to the outliers which carry weak innovation. Moreover, the experiments with real and synthetic data demonstrated the robustness of the proposed method against the strong presence of noise.

\newpage\clearpage

\bibliography{example_paper}
\bibliographystyle{plain}

\newpage\clearpage

\noindent\textbf{Appendix}

\section{Proofs of the Presented Theoretical Results}
In this section, the proofs for Theorem~\ref{theo:suffic_random2} and Theorem~\ref{theor:structured} are provided. Before we present the proofs, we review few useful lemmas adapted from~\cite{rahmani2017coherence22f,park2014greedy,ledoux2005concentration,milman2009asymptotic,hoeffding1963probability,hystack}.

\begin{lemma}
\cite{park2014greedy,ledoux2005concentration,milman2009asymptotic}
Let the columns of $\bF \in \mathbb{R}^{N \times {r}}$ be an orthonormal basis for an ${r}$-dimensional random subspace drawn uniformly at random in an ambient $N$-dimensional space. For a unit $\ell_2$-norm vector $\bc \in \mathbb{R}^{N \times 1}$
\begin{eqnarray*}
\mathbb{P} \left[ \| \bc^T \bF \|_2 > \sqrt{\frac{{ c_{\delta} r}}{N}}  \right] \leq \delta \:,
\end{eqnarray*}
where $\sqrt{c_{\delta}} = 3 \max \left(1 ,  \sqrt{\frac{8  N \pi }{(N - 1)r}} , \sqrt{\frac{8 N \log1/\delta }{(N - 1)r}} \right) $.
\label{lm:projectranodm}
\end{lemma}

\begin{lemma}
\cite{hoeffding1963probability}
For a given vector $\ba \in \mathbb{R}^{M_1 \times 1}$
\begin{eqnarray*}
\mathbb{P} \left[ \left| \sum_{i}  \epsilon_i \: \ba(i) \right| > t \|\ba \|_2 \right] < \exp(-t^2) \:,
\end{eqnarray*}
where $\{\epsilon_i \}_{i=1}^{M_1}$ are independent Rademacher random variables.
\label{lm:ramche}
\end{lemma}

\begin{lemma}~\cite{hystack}
Suppose $\bg_1, ... ,\bg_{n}$   are i.i.d. random vectors distributed uniformly on the unit sphere $\mathbb{S}^{N - 1}$ in $\mathbb{R}^{N  }$. If $N > 2$, then
\begin{eqnarray*}
\underset{\|\bu\| = 1}{\inf} \:\: \sum_{i = 1}^{n} | \bu^T \bg_i | > \sqrt{\frac{2}{\pi}} \frac{n}{\sqrt{N}} - 2\sqrt{n} - \sqrt{\frac{2 n \log \frac{1}{\delta}}{N -1 }}
\end{eqnarray*}
with probability at least $1 - \delta$.
\label{lm:perm_negative}
\end{lemma}

\begin{lemma}
\cite{rahmani2017coherence22f}
Suppose $\bg_1, ... ,\bg_{n}$   are i.i.d. random vectors distributed uniformly on the unit sphere $\mathbb{S}^{N - 1}$ in $\mathbb{R}^{N  }$. If $N > 2$, then
\begin{eqnarray}\notag
\underset{\|\bu\| = 1}{\sup} \:\: \sum_{i = 1}^{n} | \bu^T \bg_i | <  \frac{n}{\sqrt{N}} + 2\sqrt{n} + \sqrt{\frac{2 n \log \frac{1}{\delta}}{N -1 }}
%\label{suff_ng_perm}
\end{eqnarray}
with probability at least $1 - \delta$.
\label{lm:perm_positive}
\end{lemma}

\begin{lemma}
\cite{rahmani2017coherence22f}
Suppose $\bg_1, ... ,\bg_{n}$   are i.i.d. random vectors distributed uniformly on the unit sphere $\mathbb{S}^{N - 1}$ in $\mathbb{R}^{N  }$. If $N > 2$, then
\begin{eqnarray}\notag
\underset{\|\bu\| = 1}{\sup} \:\: \sum_{i = 1}^{n} (\bu^T \bg_i)^2 \leq \frac{n}{N} + \eta
%\label{eq:suff_perm_2}
\end{eqnarray}
with probability at least $1 - \delta$ where  $ \eta = \max \left( \frac{4}{3} \log \frac{2N}{\delta} , \sqrt{4 \frac{n}{N} \log \frac{2 N}{\delta}} \right) $.
\label{lm:mylemma}
\end{lemma}

\begin{lemma}
Suppose $\bF \in \mathbb{R}^{N \times r}$ spans a random $r$-dimensional  subspace. For a given vector $\bc \in \mathbb{R}^{N \times 1}$
\begin{eqnarray*}
\mathbb{P} \left[ \| \bc^T \bF \|_2 > \sqrt{\frac{c_1 \Bar{r}}{N}}  \right] \leq 1 - c_2 N^{-3} \log N  \:,
\end{eqnarray*}
where $c_1$ and $c_2$ are constant real numbers and $\Bar{r} = \max (r , \log N)$.
\end{lemma}

\begin{lemma}
Suppose $\bg_1, ... ,\bg_{n}$   are i.i.d. random vectors distributed uniformly on the unit sphere $\mathbb{S}^{N - 1}$ in $\mathbb{R}^{N  }$ and define $\mu_N = \sqrt{2} \frac{\Gamma( (N+1)/2 )}{N/2}$ where $\Gamma(\cdot)$ is the Gamma function~\cite{davis1959leonhard}. If $N > 2$, then
\begin{eqnarray*}
\mathbb{P} \left[ \left| \sum_{i = 1}^{n} | \bu^T \bg_i | - \frac{n}{\mu_N } \sqrt{\frac{2}{\pi}} \right| > 2\sqrt{n} + \sqrt{\frac{2 n \log \frac{1}{\delta}}{N -1 }} \right] < \delta \:.
\end{eqnarray*}
\label{lm:mean_exact}
\end{lemma}

\subsection{Proof of Theorem~\ref{theo:suffic_random2}}
\label{sec:_proof_1}
Theorem~\ref{theo:suffic_random2} is proved in two steps. First we provide the sufficient conditions to guarantee that when $\bd$ is an outlier, the optimal point of (\ref{opt:asli1}) is highly incoherent with $\calU$ with high probability. Second, it is shown that if $n_i/r$ is sufficiently large and $\bc^{*}$ is close to $\calU^{\perp}$, then (\ref{cond:Innovation_condition}) holds with high probability. The algorithm does not require the optimal vectors $\{ \bc_i^{*} \}_{i=1}^{n_o}$ to be orthogonal to $\calU$. As long as they are sufficiently close to $\calU^{\perp}$, iSearch works well. However, in order to simplify the analysis, in the first step of the proof we derive the sufficient conditions to guarantee that the optimal vectors  $\{ \bc_i^{*} \}_{i=1}^{n_o}$ lie in $\calU^{\perp}$.
\noindent
 The following lemma provides the sufficient conditions to guarantee that
\begin{eqnarray}
\underset{ \bc \in \calU^{\perp} \atop \bd^T\bc = 1}{\arg\min} \: \:  \| \bc^T \bD \|_1 = \underset{ \bd^T \bc = 1}{\arg\min} \: \:  \| \bc^T \bD \|_1 \:.
\label{eq:suff_hgjhd}
\end{eqnarray}
Note that in this Lemma we do not make any assumption about the distribution of the inliers and the distribution of the outliers.
Lemma~\ref{lm:main_intr} only uses Data Model 1 which is a deterministic model.

\begin{lemma}
Suppose $\bD$ follows Data Model 1. Define $\calI_0 = \{ i \in [M_2] : \hat{\bc}^T \bd_i = 0,\: \bd_i \in \bB \}$, where $\hat{\bc}$ is the optimal point of
\begin{eqnarray}
\underset{ \bc}{\min} \: \:  \| \bc^T \bD \|_1 \quad \text{subject to} \qquad \bc^T \bd = 1 \quad \text{and} \quad \bc \in \calU^{\perp} \:.
\label{opt:jaali}
\end{eqnarray}
Suppose $\bd$ is one of the columns of $\bB$ and define $\bd^{\perp}  = (\bI - \bU\bU^T)\bd/ \| (I - \bU\bU^T)\bd/\|_2$. If
\begin{eqnarray}
\begin{aligned}
& \frac{1}{2} \underset{\delta \in \calU  \atop \| \delta \|_2 = 1}{\inf} \sum_{\bd_i \in \bA} \left| \mathbf{\delta}^T \bd_i \right|  >  \underset{\delta \in \calU  \atop \| \delta \| = 1}{\sup} \sum_{\bd_i \in \bB \atop i \in \calI_0}  \left| \delta^T \bd_i \right| +  \| \bU^T \bo \|_2 \\
& \underset{\delta \in \calU \atop \| \delta \|_2 = 1}{\inf} \: \: \sum_{\bd_i \in \bA} \left| \mathbf{\delta}^T \bd_i \right|  > \\
& \quad \quad \quad \frac{2 \| \bd^T \bU \|_2}{\sqrt{1 - \|\bd^T \bU \|_2^2}} \left( \bo^T \bd^{\perp} +   \sum_{\bd_i \in \bB \atop i \in \calI_0}  \left|  \bd_i^T \bd^{\perp}  \right| \right) \: ,
\end{aligned}
\label{suf:mainone}
\end{eqnarray}
then the equality (\ref{eq:suff_hgjhd}) holds
where $\bo = \sum_{\bd_i \in \bB} \sgn (\hat{\bc}^T \bd_i) \:  \bd_i$.
\label{lm:main_intr}
\end{lemma}

In Theorem~\ref{theo:suffic_random2}, the distribution of the inliers and the distribution of the outliers follow Assumption~\ref{assum_DistUni}. The following lemma shows that if $n_i/r$ is sufficiently large, the sufficient conditions of Lemma~\ref{lm:main_intr} are satisfied with high probability.

\begin{lemma}
Suppose $\bD$ follows Assumption~\ref{assum_DistUni}. If
\begin{eqnarray}
\begin{aligned}
& \sqrt{\frac{2}{\pi}} \frac{n_i}{\sqrt{r}} - 2\sqrt{n_i} - \sqrt{\frac{2 n_i \log \frac{1}{\delta}}{r -1 }}\\
>& 2 n_z^{'} \sqrt{\frac{c_\delta r}{M_1}}+2 \sqrt{ \frac{n_o  c_\delta r \log n_o/\delta}{M_1}}
\end{aligned}
\label{suf:avalirandom}
\end{eqnarray}
and
\begin{eqnarray}
\begin{aligned}
& \sqrt{\frac{2}{\pi}} \frac{n_i}{{r}} - 2\sqrt{n_i/r} - \sqrt{\frac{2 n_i \log \frac{1}{\delta}}{(r -1)r }} > \Bigg[ \frac{n_o }{{M_1}} + 2\sqrt{\frac{n_o }{M_1}} +
\\
& \sqrt{\frac{2 n_o \log 1/\delta}{(M_1 - 1)M_1}} + \sqrt{\frac{n_o c_{\delta}^{''} \log n_o/\delta }{M_1^2 }} +  n_z^{'} \sqrt{\frac{c_{\delta}^{''}}{M_1^2}} +
\\
& \sqrt{ \left( \frac{n_o}{M_1^2}+ \frac{\eta_{\delta}}{M_1} \right) \log n_o/\delta } \Bigg] \sqrt{\frac{ 4 M_1 c_{\delta}}{M_1 - c_{\delta} r}}
\end{aligned}
\label{eq:second_goy}
\end{eqnarray}
then the optimal point of (\ref{opt:asli1}) is equal to the optimal point of (\ref{opt:jaali})
for all $\{\bd = \bd_i \: : \: \bd_i \in \bB \}$ with probability at least $1 - 7 \delta$ where $\sqrt{c_{\delta}} = 3 \max \left(1 ,  \sqrt{\frac{8  M_1 \pi }{(M_1 - 1)r}} , \sqrt{\frac{8 M_1 \log n_0/\delta }{(M_1 - 1)r}} \right) $, $\sqrt{c_{\delta}''} = 3 \max \left(1 ,  \sqrt{\frac{8  M_1 \pi }{M_1 - 1}} , \sqrt{\frac{16 M_1 \log n_0/\delta }{M_1 - 1}} \right)$, and $ \eta_{\delta} = \max \left( \frac{4}{3} \log \frac{2M_1}{\delta} , \sqrt{4 \frac{n_o}{M_1} \log \frac{2 M_1}{\delta}} \right) $.
\label{lm:suffic_random}
\end{lemma}

The following Lemma assumes that the vectors $\{ \bc_i^{*} \}_{i=1}^{n_o}$ lie in $\calU^{\perp}$ and provides the sufficient condition  to guarantee that the Innovation Values corresponding to the outliers are greater than the Innovation Values corresponding to the inliers.

\begin{lemma}
Suppose $\bD$ follows Assumption~\ref{assum_DistUni}. If the optimal vectors $\{ \bc_i^{*} \}_{i=1}^{n_o}$ lie in $\calU^{\perp}$ and
\begin{eqnarray}
\begin{aligned}
& \sqrt{\frac{2}{\pi}} \frac{n_i}{\sqrt{r}} - 2\sqrt{n_i} - \sqrt{\frac{2 n_i \log \frac{1}{\delta}}{r -1 }} >\\
& \left( \chi  - \sqrt{\frac{2}{\pi}}  \right)
 \frac{n_o}{\sqrt{M_1}} +
 2\sqrt{n_o} (1 + \sqrt{\chi }) +2 \sqrt{\frac{2 \chi  n_o \log \frac{1}{\delta}}{M -1 }}
\end{aligned}
\label{eq:akahar}
\end{eqnarray}
then $$\max \left(  \{ 1/ \| \bD^T \bc_i^{*} \|_1 \}_{i=n_o+1}^{M_2}  \right) < \min \left(  \{ 1/\| \bD^T \bc_j^{*} \|_1 \}_{j=1}^{n_o}  \right)$$ with probability at least $1 - 3\delta$.
\label{lm:intermediate_last}
\end{lemma}

\noindent
According to  Lemma~\ref{lm:suffic_random} and  Lemma~\ref{lm:intermediate_last} and according to the analysis presented in the Proof of Lemma~\ref{lm:suffic_random} and Lemma~\ref{lm:intermediate_last}, if the sufficient conditions of Theorem~\ref{theo:suffic_random2} are satisfied then the inequality (\ref{cond:Innovation_condition})  holds and $\calU$ is recovered exactly with probability \textcolor{black}{ at least $1 - 7\delta$}.

\subsection{Proof of Theorem~\ref{theor:structured}}
The iSearch method does not require the optimal vectors $\{ \bc_i^{*} \}_{i=1}^{n_o}$ to be orthogonal to $\calU$. If they are sufficiently close to $\calU^{\perp}$, iSearch works well. However, in order to simplify the analysis, in the first step of the proof we derive the sufficient conditions to guarantee that the optimal vectors $\{ \bc_i^{*} \}_{i=1}^{n_o}$ lie in $\calU^{\perp}$.

\begin{lemma}
Suppose the distribution of the inliers follows Assumption~\ref{assum_DistUni}, the distribution of the outliers  follows Assumption~\ref{asm:clus}, $\eta < | \bq^T \bq^{\perp} |$, and  $\calQ$ is equal to the column-space of $[\bU \:\: \bq]$. Define $\bq^{\perp} = \frac{(\bI -\bU \bU^T)\bq}{ \| (\bI -\bU \bU^T)\bq \|_2} $ and define $\beta = \max \left( \{ \frac{ \| \bd_i^T \bU \|_2}{ | \bd_i^T \bq^{\perp} | } \: : \: \bd_i \in \bB \} \right)$. If
\begin{eqnarray}
\begin{aligned}
& \sqrt{\frac{2}{\pi}} \frac{n_i}{{\sqrt{r}}} - 2\sqrt{n_i} - \sqrt{\frac{2 n_i \log \frac{1}{\delta}}{r -1 }}\\
>& \frac{2}{\sqrt{1 + \eta^2}} \left( n_o \|\bU^T \bq \|_2 + \eta  \sqrt{\frac{n_o r c_{\delta} \log n_o/\delta}{M_1} } \right)
\end{aligned}
\label{eq:sf11}
\end{eqnarray}
and
\begin{eqnarray}
\begin{aligned}
& \sqrt{\frac{2}{\pi}} \frac{n_i}{\sqrt{r}} - 2\sqrt{n_i} - \sqrt{\frac{2 n_i \log \frac{1}{\delta}}{r -1 }}\\>
& \frac{2 \beta}{\sqrt{1+\eta^2}} \Big( n_o \|\bq^T \bq^{\perp} \|_2 + \eta  \sqrt{\frac{n_o  c_{\delta}'' \log n_o/\delta}{M_1} } \Big)
\end{aligned}
\label{eq:sf112}
\end{eqnarray}
then the optimal point of (\ref{opt:withspan}) is equal to $\frac{\bq^{\perp}}{\bd^T \bq^{\perp}}$ for all $\{\bd = \bd_i \: : \: \bd_i \in \bB\}$ with probability at least $1 - 5\delta$ where  $\sqrt{c_{\delta}''} = 3 \max \left(1 ,  \sqrt{\frac{8  M_1 \pi }{M_1 - 1}} , \sqrt{\frac{8 M_1 \log n_0/\delta }{M_1 - 1}} \right)$.
\label{lm:orth_clust}
\end{lemma}

\noindent
The following lemma shows that if $n_i/r$ is sufficiently large and the optimal directions corresponding to the outliers are aligned with $\bq^{\perp}$, then (\ref{cond:Innovation_condition})  holds with high probability.

\begin{lemma}
Suppose the distribution of the data is similar to the presumed data distribution in Lemma~\ref{lm:orth_clust} and define $\beta = \max \left( \{ \frac{1}{| \bd_i^T \bq^{\perp}|}  \: : \: \bd_i \in \bB \} \right)$. If the optimal point of (\ref{opt:withspan}) is equal to $\frac{1}{  \bd^T \bq^{\perp}  } \bq^{\perp}  $ for all $\{ \bd = \bd_i \: : \: \bd_i \in \bB \}$ and if
\begin{eqnarray}
\begin{aligned}
& \sqrt{\frac{2}{\pi}} \frac{n_i}{\sqrt{r}} - 2\sqrt{n_i} - \sqrt{\frac{2 n_i \log \frac{n_i}{\delta}}{r -1 }}\\ >
& \frac{ \beta }{\sqrt{1+\eta^2}} \left( n_o |\bq^T \bq^{\perp}| + \eta n_o \sqrt{\frac{  c_{\delta}'' \log n_o/\delta}{M_1} }  \right)
\end{aligned}
\label{eq:tabozha}
\end{eqnarray}
then  (\ref{cond:Innovation_condition})  holds with probability at least $1- 2 \delta$.
\label{lm:finallle}
\end{lemma}

\noindent
According to Lemma~\ref{lm:orth_clust}, Lemma~\ref{lm:finallle}, and   the analysis presented in the proof of these lemmas, if the sufficient conditions of Theorem~\ref{theor:structured} are satisfied, iSearch yields exact subspace recovery with probability at least $1- 5\delta$.

\subsection{Proof of Theorem~\ref{theo:linear_dep}}
The procedure we use to prove Theorem~\ref{theo:linear_dep} is similar to procedure we employed to  prove  Theorem~\ref{theo:suffic_random2} in Section~\ref{sec:_proof_1}.
 First we provide the sufficient conditions to guarantee that when $\bd$ is an outlier, the optimal point of (\ref{opt:asli1}) is highly incoherent with $\calU$. Second, it is shown that if $n_i/r$ is sufficiently large and $\bc^{*}$ is close to $\calU^{\perp}$, then (\ref{cond:Innovation_condition}) holds. The algorithm does not require the optimal vectors $\{ \bc_i^{*} \}_{i=1}^{n_o}$ to be orthogonal to $\calU$. As long as they are sufficiently close to $\calU^{\perp}$, iSearch works well. However, in order to simplify the analysis, in the first step of the proof we derive the sufficient conditions to guarantee that the optimal vectors  $\{ \bc_i^{*} \}_{i=1}^{n_o}$ lie in $\calU^{\perp}$.

\begin{lemma}
Suppose the distribution of the inliers/outliers follows Assumption-\ref{assum_DistUni}/Assumption-\ref{asm:out}.
If
\begin{eqnarray}
\begin{aligned}
& \sqrt{\frac{2}{\pi}} \frac{n_i}{\sqrt{r}} - 2\sqrt{n_i} - \sqrt{\frac{2 n_i \log \frac{1}{\delta}}{r -1 }}\\ >
&2 n_z^{'} \| \bU^T \bU_o \| + 2 \| \bU^T \bU_o \| \sqrt{n_o \log n_o/\delta }
\end{aligned}
\label{eq:suff_1_orth_dep}
\end{eqnarray}
and
\begin{eqnarray}
\begin{aligned}
&  \sqrt{\frac{2}{\pi}} \frac{n_i}{\sqrt{r}} - 2\sqrt{n_i} - \sqrt{\frac{2 n_i \log \frac{1}{\delta}}{r -1 }}  > \frac{2  \| \bU^T \bU_o \| }{\xi}\Bigg( \frac{n_o }{\sqrt{r_o}} + \\
&2\sqrt{n_o } +  \sqrt{\frac{2 n_o \log \frac{1}{\delta}}{r_o -1 }} + 2\sqrt{  \left( \frac{n_o}{r_o} + \eta_{\delta}^{'} \right)\log \frac{n_o}{\delta}  }  +n_z^{'} \Bigg)
\end{aligned}
\label{eq:suff_2_orth_dep}
\end{eqnarray}
then the optimal vectors $\{ \bc^{*}_i \}_{i=1}^{n_o}$ lie in $\calU^{\perp}$ with probability at least $1 - 5\delta$ where $\eta_{\delta}^{'} = \max \left(\frac{4}{3} \log 2 r_o /\delta \: , \: \sqrt{4 \frac{n_o}{r_o} \log \frac{2 r_d}{\delta}} \right) $ and $\xi = \frac{\min \left( \left\{ \|\bb_j^T \bU^{\perp} \|_2 \right\}_{j=1}^{n_o} \right)}{\|\bU_o^T \bU^{\perp} \|}$.
\label{lm:orh_sb_clus}
\end{lemma}

Next Lemma assumes that the optimal vectors $\{ \bc^{*}_i \}_{i=1}^{n_o}$ lie in $\calU^{\perp}$ and shows that if the number of inliers is sufficiently larger than the number of outliers, the inequality (\ref{cond:Innovation_condition}) holds with high probability.

\begin{lemma}
Suppose the distribution of the inliers/outliers follows Assumption-\ref{assum_DistUni}/Assumption-\ref{asm:out} and suppose the optimal vectors $\{ \bc^{*}_i \}_{i=1}^{n_o}$ lie in $\calU^{\perp}$. If
\begin{eqnarray}
\begin{aligned}
& \sqrt{\frac{2}{\pi}} \frac{n_i}{\sqrt{r}} - 2\sqrt{n_i} - \sqrt{\frac{2 n_i \log \frac{1}{\delta}}{r -1 }}\\ >
&\|\bU_o^T \bU^{\perp} \|_2 \Bigg(  \frac{ \chi n_o}{\sqrt{r_o}} + 2\sqrt{ \chi n_o} +\sqrt{ \chi \frac{2 n_o \log \frac{1}{\delta}}{r_o -1 }} \Bigg)
\end{aligned}
\label{eq:valueinlemmasub}
\end{eqnarray}
then the inequality (\ref{cond:Innovation_condition})  holds with probability at least $1 - 2\delta$.
\label{lm:values_sub_lin_dep}
\end{lemma}

According to Lemma~\ref{lm:orh_sb_clus}, Lemma~\ref{lm:values_sub_lin_dep}, and the analysis presented in the proof of these lemmas, if the sufficient conditions of Theorem~\ref{theo:linear_dep} are satisfied,
then  (\ref{cond:Innovation_condition})  holds and $\calU$ is recovered exactly with probability at least $1- 5 \delta$.

\subsection{Proof of Theorem~\ref{theo:suffic_random24}}
\textcolor{black}{
The only difference between Theorem~\ref{theo:suffic_random2} and Theorem~\ref{theo:suffic_random24} is the difference between the presumed distribution of the inliers.} In order to derive the sufficient conditions, we only need to establish a new lower-bound for $\underset{\delta \in \calU  \atop \| \delta \| = 1}{\inf} \sum_{\bd_i \in \bA} | \delta^T \bd_i |$.
According to the clustering structure of the inliers,
\begin{eqnarray}
 \begin{aligned}
& \underset{\delta \in \calU  \atop \| \delta \| = 1}{\inf} \sum_{\bd_i \in \bA} \left| \mathbf{\delta}^T \bd_i \right| \\\ge
& \underset{\delta \in \calU  \atop \| \delta \| = 1}{\inf} \sum_{k=1}^{m} \| \delta^T \bU_k \|_2 \left(  \underset{\delta_k \in \calU_k  \atop \| \delta \| = 1}{\inf}  \| \delta_k^T \bA_k \|_1  \right) \:.
\end{aligned}
\label{eq:jdd1}
\end{eqnarray}

\noindent
Therefore, according to (\ref{eq:jdd1}) and the definition of integer $g$,
 \begin{eqnarray*}
 \begin{aligned}
& \underset{\delta \in \calU  \atop \| \delta \| = 1}{\inf} \sum_{\bd_i \in \bA} \left| \mathbf{\delta}^T \bd_i \right| \\\ge
&\left(  \underset{\delta \in \calU_g  \atop \| \delta \| = 1}{\inf}  \| \delta^T \bA_g \|_1  \right) \underset{\delta \in \calU  \atop \| \delta \| = 1}{\inf} \sum_{k=1}^{m} \| \delta^T \bU_k \|_2  \:.
\end{aligned}
\end{eqnarray*}
Parameter $\rho$ was defined as $\rho = \underset{\delta \in \calU  \atop \| \delta \| = 1}{\inf} \sum_{k=1}^{m} \| \delta^T \calU_k \|_2$. The parameter $\rho$ shows how well the columns of $\bA$ are distributed in $\calU$. According to the definition of $\rho$ and using Lemma~\ref{lm:perm_negative}, (\ref{eq:jdd1}) can be upper-bounded as follows
 \begin{eqnarray*}
 \begin{aligned}
& \underset{\delta \in \calU  \atop \| \delta \| = 1}{\inf} \sum_{\bd_i \in \bA} \left| \mathbf{\delta}^T \bd_i \right| \\\ge
&\rho \left( \sqrt{\frac{2}{\pi}} \frac{n_g}{\sqrt{r}} - 2\sqrt{n_g} - \sqrt{\frac{2 n_g \log \frac{1}{\delta}}{r -1 }}  \right) \:.
\end{aligned}
\end{eqnarray*}

\subsection{Proof of Lemma~\ref{them:stability}}

The vector $\bc^{*}$ was defined as the optimal direction. If $\bd$ is an outlier, then by definition $\bd$ has non-zero projection on $\calU^{\perp}$. Define $\bR$ as an orthonormal basis for $\calU^{\perp}$ and define $\bd^{\perp} = \frac{\bR \bR^T \bd}{\| \bd^T \bR \|_2^2}$. Accordingly to the definition of $\bd^{\perp}$, $\bd^T \bd^{\perp} = 1$ which implies that when $\bd$ is an outlier we can conclude that
%and we define $\bc^{'} =\frac{ \| \bc^{*}\|_2}{ \| (\bI - \bU \bU^T)\bc^{*} \|_2} (\bI - \bU \bU^T)\bc^{*} $ which means that $\bc^{'} \in \calU^{\perp}$.
%Define vector $\bc^{*}$  as the optimal direction of (\ref{opt:asli1}) when $\sigma_n =0$ and define  $\bc_e^{*}$ as  the optimal direction of (\ref{opt:asli1}) when $\sigma_n$ is non-zero (inliers are noisy).
%According to Lemma~\ref{lm:mean_exact}, if $\bc^{*} \in \calU^{}\perp$, then
\begin{eqnarray*}
\begin{aligned}
& \| \bD^T \bc^{*} \|_1 \le \| \bD^T \bd^{\perp}  \|_1 = \| \bB^T \bd^{\perp} \|_1 + \| \bE^T \bd^{\perp} \|_1 \\
\le& \frac{1}{\| \bd^T \bR \|_2} \left( \frac{n_o}{\mu_{M_1} } \sqrt{\frac{2}{\pi}} + 2\sqrt{n_o} + \sqrt{\frac{2 n_o \log \frac{1}{\delta}}{M_1 -1 }} \right) +\\
& \frac{\sigma_n}{(1+\sigma_n^2) \| \bd^T \bR \|_2} \left( \frac{n_i}{\mu_{M_1} } \sqrt{\frac{2}{\pi}} + 2\sqrt{n_i} + \sqrt{\frac{2 n_i \log \frac{1}{\delta}}{M_1 -1 }} \right) \:.
\end{aligned}
\label{eq:ineq_jadid}
\end{eqnarray*}
with probability at least $1 - 2\delta$.
\noindent
Next, we can derive the following lower-bound
\begin{eqnarray}\notag
\begin{aligned}
& \|  \bD^T \bc^{*} \|_1 \ge \frac{1}{1+ \sigma_n^2}  \left(\|  \bA^T \bc^{*} \|_1 -  \| \bE^T \bc^{*} \|_1\right) +  \|  \bB^T \bc^{*} \|_1 \\
\ge& \frac{\varrho}{1+\sigma_n^2} \left(  \frac{ \: n_i}{ \sqrt{r} } \sqrt{\frac{2}{\pi}} - 2\sqrt{ \: n_i} - \sqrt{\frac{2  \: n_i \log \frac{1}{\delta}}{r -1 }} \right) + \\
& \sqrt{\frac{2}{\pi}} \frac{n_o}{ \mu_{M_1} } - 2\sqrt{n_o} - \sqrt{\frac{2 n_o \log \frac{1}{\delta}}{M_1 -1 }} -  \frac{\sigma_n}{1+\sigma_n^2}  \Bigg( \sqrt{\frac{2}{\pi}} \frac{  n_i}{ \mu_{M_1} } \\
& + 2\sqrt{ n_i} + \sqrt{\frac{2 n_i \log \frac{1}{\delta}}{M_1 -1 }} \Bigg) \:,
 %\label{eq:neww2}
 \end{aligned}
\end{eqnarray}
which is valid with probability at least $1 - 3 \delta$.
%Since $\bc_e^{*}$ is the optimal direction for the noisy data, $\| \bD^T \bc_e^{*}\|_1 \le \| \bD^T \bc^{*}\|_1 $ when $\sigma_n$ is non-zero.
The value of the lower-bound is smaller than the value of the upper-bound  and $ \sqrt{M_1 - 0.5}< \mu_{M_1}$~\cite{lerman2015robustnn}. Accordingly, we can conclude that
(\ref{eq:bounded_rho}) holds with probability at least $1 - 4\delta$.

\subsection{Proof of Theorem~\ref{thm:final}}
Suppose $\bd$ in (\ref{opt:asli1}) is an inlier and $\bd = \frac{1}{1 + \sigma_n^2}(\ba + \be)$ where $\ba$ is a clean inlier and $\be$ is the added noise. Since $ \bd^T \bc^{*} = 1$,  then according to the presumed noisy data model, $\| \bU^T \bc^{*} \|_2 \ge (1 + \sigma_n^2)/(1 + \sigma_n)$.
In addition,  we can conclude that $\| \bc^{*} \|_2 \ge (1+ \sigma_n^2)/(1 + \sigma_n)$.
Therefore,
\begin{eqnarray*}
\begin{aligned}
 \| \bD^T \bc^{*} \|_1 & = \frac{1}{1 + \sigma_n^2}\|(\bA + \bE)^T \bc^{*} \|_1 + \|\bB^T \bc^{*} \|_1 \\
& \ge \frac{1}{1 + \sigma_n} \underset{\delta \in \calU  \atop \| \delta \|_2 = 1}{\inf} \| \delta^T \bA \|_1 + \frac{1}{1 + \sigma_n} \underset{ \| \delta \|_2 = 1}{\inf}  \| \delta^T \bB \|_1 \\
  & \hspace{2cm} -\chi^{'} \underset{ \| \delta \|_2 = 1}{\sup}  \| \delta^T \bE \|_1 \\
 & \ge \frac{1}{1 + \sigma_n} \Bigg( \sqrt{\frac{2}{\pi}} \frac{n_i}{{\mu_r}} - 2\sqrt{n_i} - \sqrt{\frac{2 n_i \log \frac{1}{\delta}}{r -1 }} \Bigg) \\
& \hspace{.1cm }+ \frac{1}{1 + \sigma_n} \Bigg( \sqrt{\frac{2}{\pi}} \frac{n_o}{\mu_{M_1}}  - 2\sqrt{n_o} - \sqrt{\frac{2 n_o \log \frac{1}{\delta}}{M_1 -1 }}  \Bigg) \\
&  \hspace{.1cm } - \sigma_n \chi^{'} \Bigg( \sqrt{\frac{2}{\pi}} \frac{ n_i}{\mu_{M_1}}
 + 2\sqrt{n_i} + \sqrt{\frac{2  n_i \log \frac{1}{\delta}}{M_1 -1 }} \Bigg)
\end{aligned}
\end{eqnarray*}
with probability at least $1-3\delta$.

If $\bd$ is an outlier, then
similar to the proof of Lemma~\ref{them:stability},
\begin{eqnarray*}
\begin{aligned}
& \| \bD^T \bc^{*} \|_1 \le \| \bD^T \bd^{\perp}  \|_1 = \| \bB^T \bd^{\perp} \|_1 + \frac{1}{1 + \sigma_n^2}\| \bE^T \bd^{\perp} \|_1 \\
\le& \frac{1}{\| \bd^T \bR \|_2} \left( \frac{n_o}{\mu_{M_1} } \sqrt{\frac{2}{\pi}} + 2\sqrt{n_o} + \sqrt{\frac{2 n_o \log \frac{1}{\delta}}{M_1 -1 }} \right) +\\
& \frac{\sigma_n}{(1+\sigma_n^2) \| \bd^T \bR \|_2} \left( \frac{n_i}{\mu_{M_1} } \sqrt{\frac{2}{\pi}} + 2\sqrt{n_i} + \sqrt{\frac{2 n_i \log \frac{1}{\delta}}{M_1 -1 }} \right) \:.
\end{aligned}
\label{eq:ineq_jadid}
\end{eqnarray*}
with probability at least $1 - 2\delta$. Accordingly, if the sufficient conditions of Theorem~\ref{thm:final} are satisfied,  Innovation Values of the outliers are larger than the Innovation Values of the inliers with probability at least $1 - 5\delta$.

\section{Proofs of the Intermediate Results}

\noindent
\textbf{Proof of Lemma~\ref{lm:main_intr}}\\
The optimization problem (\ref{opt:asli1}) is a convex optimization problem. In order to prove that the equality (\ref{eq:suff_hgjhd}) holds,
it is enough to show that the cost function of (\ref{opt:asli1}) increases if we move away from $\hat{\bc}$ along any feasible direction where $\hat{\bc}$ is the optimal point of (\ref{opt:jaali}). Define $$\omega (\delta) =   \| ( \hat{\bc} - \mathbf{\delta})^T \bD \|_1 - \| \hat{\bc}^T \bD \|_1  \: .$$
Therefore, we only need to prove that $\omega(\delta) \ge 0$ for any sufficiently small $\delta$ such that $\delta^T \bd = 0$. The deviation vector $\delta$ should be orthogonal to $\bd$ to ensure that $\hat{\bc} - \delta$ stays in the feasible set of (\ref{opt:asli1}).
%The reason that we limit the deviation vector to vectors such that $\delta^T \bd = 0$ is that the deviation $\delta$ should stay in the feasible set of ().

The vector $\hat{\bc}$ is the optimal point of (\ref{opt:jaali}). Therefore, for any vector $\delta_o$ such that
\begin{eqnarray}
\mathbf{\delta}_o^T \bd = 0 \: \: , \: \: \mathbf{\delta}_o \in \calU^{\perp}.
\label{condition_oracel1}
\end{eqnarray}
we have
\begin{eqnarray}
\| (\hat{\bc} - \mathbf{\delta}_o)^T \bD \|_1 - \| \hat{\bc}^T \bD \|_1 \ge 0 \: .
\label{strict_oracel}
\end{eqnarray}

\noindent
The vectors $\hat{\bc}$ and $\delta_o$ are orthogonal to $\calU$. Thus,  the inequality (\ref{strict_oracel}) is equivalent to
\begin{eqnarray}
\| (\hat{\bc} - \mathbf{\delta}_o)^T \bB \|_1 - \| \hat{\bc}^T \bB \|_1 \ge 0 \: .
\label{eq:new_eq}
\end{eqnarray}
When $\delta_o \to 0$, we can rewrite (\ref{eq:new_eq}) as
\begin{align}
%\begin{aligned}
& \| (\hat{\bc} - \mathbf{\delta}_o)^T \bB \|_1 - \| \hat{\bc}^T \bB \|_1 \nonumber\\
=& \sum_{\bd_i \in \bB} \left[ (\hat{\bc} - \mathbf{\delta}_o)^T \bd_i)^2 \right]^{1/2} - \sum_{\bd_i \in \bB} \left| \hat{\bc}^T \bd_i \right| \nonumber \\
=&  \sum_{\bd_i \in \bB} \hspace{-.2cm}\left[ (\hat{\bc}^T \bd_i)^2 \hspace{-.08cm} - \hspace{-.08cm}  2 (\hat{\bc}^T \bd_i)(\mathbf{\delta}_o^T \bd_i) \hspace{-.1cm}+\hspace{-.1cm} (\mathbf{\delta}_o^T \bd_i)^2 \right]^{1/2} \hspace{-.15cm} - \hspace{-.2cm} \sum_{\bd_i \in \bB} \left| \hat{\bc}^T \bd_i \right| \nonumber\\
=&  \sum_{\bd_i \in \bB \atop i \in \calI_0} \left| \delta_o^T \bd_i \right|
+ \sum_{\bd_i \in \bB \atop i \in \calI^c_0} \left|\hat{\bc}^T \bd_i \right| \bigg[ 1 - 2 \frac{\sgn (\hat{\bc}^T \bd_i)}{|\hat{\bc}^T \bd_i|} (\mathbf{\delta}_o^T \bd_i) \nonumber\\
& \qquad\qquad\qquad\qquad +   \calO (\| \delta_o \|^2 )\bigg]^{1/2}  - \sum_{\bd_i \in \bB \atop i \in \calI^c_0} \left| \hat{\bc}^T \bd_i \right| \nonumber\\
=& \sum_{\bd_i \in \bB \atop i \in \calI_0} \left| \delta_o^T \bd_i \right|- \hspace{-0.25cm}\sum_{\bd_i \in \bB \atop i \in \calI^c_0} {\sgn} (\hat{\bc}^T \bd_i) (\mathbf{\delta}_o^T \bd_i) + \mathcal{O} ( \|\mathbf{\delta_o}\|^2)
%\end{aligned}
\label{eq:madar_of_exp}
\end{align}
where the last identity follows from the Taylor expansion of the square root function. Thus,
\begin{eqnarray}
\sum_{\bd_i \in \bB \atop i \in \calI_0} \left| \delta_o^T \bd_i \right| - \sum_{\bd_i \in \bB \atop i \in \calI^c_0} {\sgn} (\hat{\bc}^T \bd_i) (\mathbf{\delta}_o^T \bd_i) + \mathcal{O} ( \|\mathbf{\delta}_o\|^2)
\label{eq:komak}
\end{eqnarray}
has to be greater than zero for any small $\delta_o$~which satisfies~(\ref{condition_oracel1}).

The function $\omega(\delta)$ can be expanded as
\begin{eqnarray}
\begin{aligned}
& \omega (\delta) =   \| ( \hat{\bc} - \mathbf{\delta})^T \bA \|_1 - \| \hat{\bc}^T \bA \|_1 +  \\
& \quad \quad \quad \quad \quad \| ( \hat{\bc} - \mathbf{\delta})^T \bB \|_1 - \| \hat{\bc}^T \bB \|_1  \: .
\end{aligned}
\label{eq:expanded1}
\end{eqnarray}
We decompose $\delta$ in (\ref{eq:expanded1})
as $\delta = \delta_i + \delta_o$ where $ \delta_i$ lies in $\calU$ and $ \delta_o$ lies in $\calU^{\perp}$. Thus, $\| ( \hat{\bc} - \mathbf{\delta})^T \bA \|_1 - \| \hat{\bc}^T \bA \|_1 = \| \delta_i^T \bA \|_1$. Similar to (\ref{eq:madar_of_exp}),
as $\delta \to 0$,
\begin{eqnarray}\notag
\begin{aligned}
&\| (\hat{\bc} - \mathbf{\delta})^T \bB \|_1 - \| \hat{\bc}^T \bB \|_1 \\ =
& \sum_{\bd_i \in \bB \atop i \in \calI_0} \left| \delta^T \bd_i \right| - \sum_{\bd_i \in \bB \atop i \in \calI^c_0} \sgn (\hat{\bc}^T \bd_i) \: \delta^T \bd_i + \mathcal{O} (\| \delta^2 \|) \: .
\end{aligned}
%\label{ghesmate_2}
\end{eqnarray}
Accordingly, in order to show that (\ref{eq:suff_hgjhd}) holds, it is enough to guarantee that for any sufficiently small $\delta$ such that $\delta^T \bd = 0$,
\begin{eqnarray}
\begin{aligned}
& \omega(\delta) = \sum_{\bd_i \in \bA} \left| \mathbf{\delta}_i^T \bd_i \right|  + \sum_{\bd_i \in \bB \atop i \in \calI_0} \left| \delta^T \bd_i \right| \\
& \qquad\qquad- \sum_{\bd_i \in \bB \atop i \in \calI^c_0} \sgn (\hat{\bc}^T \bd_i) \: \delta^T \bd_i > 0 \:.
\end{aligned}
\label{eq:themain_enq}
\end{eqnarray}
%\bibliographystyle{IEEEtran}
%\bibliography{IEEEabrv,bibfile}

Let us define vector $\bo$ as
\begin{eqnarray*}
\bo = \sum_{\bd_i \in \bB} \sgn (\hat{\bc}^T \bd_i) \:  \bd_i \:.
\end{eqnarray*}
In order to show that the inequality (\ref{eq:themain_enq}) holds for any $\delta$ such that $\delta^T \bd =0$, it is sufficient to guarantee that
\begin{eqnarray}
\begin{aligned}
& \frac{1}{2} \sum_{\bd_i \in \bA} \left| \mathbf{\delta}_i^T \bd_i \right| > \sum_{\bd_i \in \bB \atop i \in \calI_0}  \left|  \delta_i^T \bd_i \right| + \delta_i^T \bo \\
& \frac{1}{2} \sum_{\bd_i \in \bA} \left| \mathbf{\delta}_i^T \bd_i \right| > \delta_o^T  \bo -  \sum_{\bd_i \in \bB \atop i \in \calI_0}  \left| \delta_o^T \bd_i \right|
\end{aligned}
\label{cond_lemman2}
\end{eqnarray}
We derive the sufficient conditions to guarantee that both of the inequalities  in (\ref{cond_lemman2}) hold. For the first inequality, it suffices to guarantee that
\begin{eqnarray}\notag
\begin{aligned}
& \frac{1}{2} \underset{\delta \in \calU  \atop \| \delta \| = 1}{\inf} \sum_{\bd_i \in \bA} \left| \mathbf{\delta}^T \bd_i \right|  >  \underset{\delta \in \calU  \atop \| \delta \| = 1}{\sup} \sum_{\bd_i \in \bB \atop i \in \calI_0}  \left| \delta^T \bd_i \right| +    \underset{\delta \in \calU  \atop \| \delta \| = 1}{\sup} \delta^T \bo
\end{aligned}
%\label{sup_min1}
\end{eqnarray}
which can be simplified to
\begin{eqnarray}
\begin{aligned}
& \frac{1}{2} \underset{\delta \in \calU  \atop \| \delta \| = 1}{\inf} \sum_{\bd_i \in \bA} \left| \mathbf{\delta}^T \bd_i \right|  >  \underset{\delta \in \calU  \atop \| \delta \| = 1}{\sup} \sum_{\bd_i \in \bB \atop i \in \calI_0}  \left| \delta^T \bd_i \right| +  \| \bU^T \bo \|_2 \:.
\end{aligned}
\label{eq:nh1}
\end{eqnarray}

For the second inequality of (\ref{cond_lemman2}), it suffices  to show that for any pair of $\delta_i$ and $\delta_o$ such that $\delta_i^T \bd = - \delta_o^T \bd$, the inequality holds. We can scale both $\delta_i$ and $\delta_o$ such that $\delta_i^T \bd = - \delta_o^T \bd = 1$. Therefore, it is enough to guarantee that
\begin{eqnarray}
\begin{aligned}
&\frac{1}{2} \underset{\delta_i \in \calU \atop \delta_i^T \bd = 1}{\inf} \: \: \sum_{\bd_i \in \bA} \left| \mathbf{\delta}_i^T \bd_i \right|\\  >
& \underset{\delta_o \in \calU^{\perp}  \atop \delta_o^T \bd = 1}{\sup} \: \: \left( \delta_o^T  \bo -  \sum_{\bd_i \in \bB \atop i \in \calI_0}  \left| \delta_o^T \bd_i \right| \right) \:.
\label{linear_cond2n2}
\end{aligned}
\end{eqnarray}

Define vector $\bd^{\perp}  = (\bI - \bU\bU^T)\bd/ \| (\bI - \bU\bU^T)\bd/\|_2$, i.e., $\bd^{\perp}$ is a unit $\ell_2$-norm vector which is aligned with the projection of $\bd$ onto $\calU^{\perp}$. Let us decompose $\delta_o$ into two components which one component is orthogonal to $\bd$.
Thus, we expand $\delta_o$ as
$
\delta_o = \delta_{oa} + \delta_{od}
$
where
\begin{eqnarray*}
\delta_{oa} = (\bI - \bd^{\perp} (\bd^{\perp})^T) \delta_o \: \: \text{and} \: \: \delta_{od} = \bd^{\perp} (\bd^{\perp})^T \delta_o \:.
\end{eqnarray*}

\noindent
In order to show that (\ref{linear_cond2n2}) holds, it suffices to ensure that the LHS of (\ref{linear_cond2n2}) is greater than
\begin{eqnarray*}
\underset{\delta_o \in \calU^{\perp}  \atop \delta_o^T \bd = 1}{\sup}  \left( \delta_{oa}^T  \bo + \delta_{od}^T  \bo -  \sum_{\bd_i \in \bB \atop i \in \calI_0}  \left| \delta_{oa}^T \bd_i \right|  +  \sum_{\bd_i \in \bB \atop i \in \calI_0}  \left| \delta_{od}^T \bd_i \right| \right)
\end{eqnarray*}
Note that
$
\delta_{oa} \in  \calU^{\perp}$ and $   \delta_{oa}^T \bd = 0 \:
$.
Therefore, according to (\ref{eq:komak}), we can conclude that $\delta_{oa}^T \bo - \sum_{\bd_i \in \bB \atop i \in \calI_0}  \left| \delta_{oa}^T \bd_i \right| \leq 0$. Therefore, it suffices to show that the LHS of (\ref{linear_cond2n2})  is greater  than
\begin{eqnarray}
\begin{aligned}
& \underset{\delta_o \in \calU^{\perp}  \atop \delta_o^T \bd = 1}{\sup}  \: \: \left( \delta_{od}^T  \bo +  \sum_{\bd_i \in \bB \atop i \in \calI_0}  \left| \delta_{od}^T \bd_i \right| \right) \\
=&  \underset{\delta_o \in \calU^{\perp}  \atop \delta_o^T \bd = 1}{\sup}  \: \: \left(   \bo^T \bd^{\perp} (\bd^{\perp})^T \delta_o +  \sum_{\bd_i \in \bB \atop i \in \calI_0}  \left|  \bd_i^T \bd^{\perp} (\bd^{\perp})^T \delta_o \right| \right) \\
=& \frac{1}{| \bd^{\perp} \bd |} \left( \bo^T \bd^{\perp} + \sum_{\bd_i \in \bB \atop i \in \calI_0}  \left|  \bd_i^T \bd^{\perp}  \right| \right)
\end{aligned}
\label{dovom_1}
\end{eqnarray}

\noindent
In addition, we can simplify the LHS of (\ref{linear_cond2n2}) as
\begin{eqnarray}
\begin{aligned}
&\frac{1}{2} \underset{\delta \in \calU \atop \delta^T \bd = 1}{\inf} \: \: \sum_{\bd_i \in \bA} \left| \mathbf{\delta}_i^T \bd_i \right|  > \frac{1}{2 \| \bd^T \bU \|_2}  \underset{\delta \in \calU \atop \| \delta \|_2 = 1}{\inf} \: \: \sum_{\bd_i \in \bA} \left| \mathbf{\delta}^T \bd_i \right|
\label{linear_cond2}
\end{aligned}
\end{eqnarray}
Therefore if (\ref{eq:nh1}) is satisfied and the RHS of (\ref{linear_cond2}) is greater than the (\ref{dovom_1}), then the equality (\ref{eq:suff_hgjhd}) holds.

\smallbreak
\noindent
\textbf{Proof of Lemma~\ref{lm:suffic_random}}\\
First we provide the sufficient conditions to guarantee that the first inequality of (\ref{suf:mainone}) holds with high probability.
The distribution of the inliers follow Assumption~\ref{assum_DistUni}. According to Lemma~\ref{lm:perm_negative},
\begin{eqnarray}
\begin{aligned}
&  \underset{\delta \in \calU  \atop \| \delta \| = 1}{\inf} \sum_{\bd_i \in \bA} \left| \mathbf{\delta}^T \bd_i \right|  > \sqrt{\frac{2}{\pi}} \frac{n_i}{\sqrt{r}} - 2\sqrt{n_i} - \sqrt{\frac{2 n_i \log \frac{1}{\delta}}{r -1 }} \:.
\end{aligned}
\label{eq:suf_1_1}
\end{eqnarray}
with probability at least $1 - \delta$. Note that (\ref{eq:suf_1_1}) does not depend on any $\bd$ in $\{ \bd = \bd_i \: : \: \bd_i \in \bB \}$.
Moreover, according to Lemma~\ref{lm:projectranodm},
\begin{eqnarray}
 \max \left( \left\{ \underset{\delta \in \calU  \atop \| \delta \| = 1}{\sup}   \left| \delta^T \bb_i \right|   \right\}_{i=1}^{n_o} \right) \le \sqrt{\frac{{ c_{\delta} r}}{M_1}}
\label{eq:new_general_bound}
\end{eqnarray}
with probability at least $1 - \delta$ where $\sqrt{c_{\delta}} = 3 \max \left(1 ,  \sqrt{\frac{8  M_1 \pi }{(M_1 - 1)r}} , \sqrt{\frac{8 M_1 \log n_0/\delta }{(M_1 - 1)r}} \right)$. Accordingly,
\begin{eqnarray}
\underset{\delta \in \calU  \atop \| \delta \| = 1}{\sup} \sum_{\bd_i \in \bB \atop i \in \calI_0}  \left| \delta^T \bd_i \right| \leq n_z \sqrt{\frac{{ c_{\delta} r}}{M_1}}
\label{eq:suf_1_2jj}
\end{eqnarray}
with probability at least $1 - \delta$ and $n_z$ is the cardinality of $\calI_0 = \{ i \in [M_2] : \hat{\bc}^T \bd_i = 0,\: \bd_i \in \bB \}$.

For an outlying column  $\bd_i$, define
\begin{eqnarray}
\begin{aligned}
& \bd_i^{\hat{\bc}} = \frac{1}{\| \hat{\bc} \|_2^2} \hat{\bc} \hat{\bc}^T \bd_i \\
& \bd_i^{\hat{\bc}^{\perp}} = \left( \bI - \frac{1}{\| \hat{\bc} \|_2^2} \hat{\bc} \hat{\bc}^T \right) \bd_i
\end{aligned}
\label{eq:dcdo}
\end{eqnarray}
Thus, we can rewrite $\bo$ as
\begin{eqnarray}
\begin{aligned}
\bo = \sum_{\bd_i \in \bB} \frac{|\hat{\bc}^T \bd_i|}{\| \hat{\bc} \|_2^2} \hat{\bc} + \sum_{\bd_i \in \bB} \sgn (\hat{\bc}^T \bd_i) \:  \bd_i^{\hat{\bc}^{\perp}} \:.
\end{aligned}
\label{eq:oooo}
\end{eqnarray}
Since the columns of $\bB$ are drawn uniformly at random  from  $\mathbb{S}^{M_1 -1 }$, the distribution of the vectors $\bb_i$ in the subspace orthogonal to $\hat{\bc}$ is independent of  $\sgn (\hat{\bc}^T \bb_i)$. Accordingly, the distribution of $\bo$ is equivalent to the distribution of $ \bo^{'} = \sum_{\bd_i \in \bB \atop i \in \calI^c_0} \frac{|\hat{\bc}^T \bd_i|}{\| \hat{\bc} \|_2^2} \hat{\bc} + \sum_{\bd_i \in \bB \atop i \in \calI^c_0}   \:   \epsilon_i \bd_i^{\hat{\bc}^{\perp}}$, where $\{ \epsilon_i\}_{i=1}^{n_o}  $ are independent Rademacher random variables.
The optimal vector
 $\hat{\bc}$ is orthogonal to $\calU$. Thus,
\begin{eqnarray}
 \underset{\delta \in \calU  \atop \| \delta \| = 1}{\sup} \delta^T \bo^{'} =  \underset{\delta \in \calU  \atop \| \delta \| = 1}{\sup} \sum_{\bd_i \in \bB \atop i \in \calI^c_0}   \:   \epsilon_i \delta^T \bd_i^{\hat{\bc}^{\perp}}
 \label{eq:taberaw}
\end{eqnarray}
Note that in (\ref{eq:taberaw}), $\delta^T \bd_i^{\hat{\bc}^{\perp}} =  \delta^T \bd_i  \le \| \bU^T \bd_i \|_2$.
Suppose (\ref{eq:new_general_bound}) is true.
We use Lemma~\ref{lm:ramche} to bound (\ref{eq:taberaw}) as
\begin{eqnarray}
\begin{aligned}
&\mathbb{P} \left[ \textcolor{black}{ \underset{\delta \in \calU  \atop \| \delta \| = 1}{\sup} }\sum_{\bd_i \in \bB \atop i \in \calI^c_0}   \:   \epsilon_i \delta^T \bd_i^{\hat{\bc}^{\perp}} \ge  \sqrt{(n - n_z)\frac{  c_\delta r \: \log n_o/\delta}{M_1}} \right] \le \\
& \quad \quad \quad \quad \quad \delta/n_o \:.
\end{aligned}
\label{eq:suf_13}
\end{eqnarray}
According to (\ref{eq:suf_1_1}), (\ref{eq:suf_1_2jj}), and (\ref{eq:suf_13}) if (\ref{suf:avalirandom}) is satisfied, the first inequality of (\ref{suf:mainone}) holds with probability at least $1-3\delta$ for all $ \{ \bd= \bb_i \}_{i=1}^{n_o}$ (because we guarantee that (\ref{eq:suff_hgjhd}) holds for all $ \{ \bd= \bb_i \}_{i=1}^{n_o}$).

In order to guarantee that the second inequality of (\ref{suf:mainone}) holds, we  need to bound ${\bo^{'}}^T \bd^{\perp}$ which can be expanded as
\begin{eqnarray}
\begin{aligned}
 {\bo^{'}}^T \bd^{\perp} =& \sum_{\bd_i \in \bB \atop i \in \calI^c_0} \frac{|\hat{\bc}^T \bd_i|}{\| \hat{\bc} \|_2^2} \hat{\bc}^T \bd^{\perp} + \sum_{\bd_i \in \bB \atop i \in \calI^c_0} \epsilon_i \: (\bd^{\perp})^T  \bd_i^{\hat{\bc}^{\perp}}\\
 \le&
   \frac{1}{\|\hat{\bc}\|_2} \sum_{\bd_i \in \bB } {|\hat{\bc}^T \bd_i|}  + \sum_{\bd_i \in \bB \atop i \in \calI^c_0} \epsilon_i (\bd^{\perp})^T  \bd_i^{\hat{\bc}^{\perp}}
\end{aligned}
\label{eq:odperp}
\end{eqnarray}
Note that
\begin{eqnarray*}
\underset{\| \delta \| = 1}{\sup} \sum_{\bd_i \in \bB} |\delta^T \bd_i| > \frac{1}{\|\hat{\bc}\|_2} \sum_{\bd_i \in \bB } {|\hat{\bc}^T \bd_i|} \:.
\end{eqnarray*}
Thus,
according to Lemma~\ref{lm:perm_positive} the first component of the RHS of (\ref{eq:odperp})  is bounded as
\begin{eqnarray}
\begin{aligned}
&\mathbb{P} \Bigg[ \frac{1}{\|\hat{\bc}\|_2} \sum_{\bd_i \in \bB } {|\hat{\bc}^T \bd_i|}  > \frac{n_o }{\sqrt{M_1}} + 2\sqrt{n_o } + \sqrt{\frac{2 n_o \log \frac{1}{\delta}}{M_1 -1 }} \Bigg] \\
& \quad \quad \quad \quad \quad \quad <\delta \:.
\end{aligned}
\label{eq:shr1}
\end{eqnarray}
The second component of the RHS of (\ref{eq:odperp}) can be expanded as
\begin{eqnarray}
\begin{aligned}
\sum_{\bd_i \in \bB \atop i \in \calI^c_0} \epsilon_i (\bd^{\perp})^T  \bd_i^{\hat{\bc}^{\perp}} = \sum_{\bd_i \in \bB \atop i \in \calI^c_0}  \left( \epsilon_i {\bd^{\perp}}^T \bd_i - \epsilon_i \frac{\hat{\bc}^T \bd^{\perp} \hat{\bc}^T \bd_i}{\| \hat{\bc} \|_2^2}  \right)
\end{aligned}
\label{eq:nnw1}
\end{eqnarray}
The vectors $\bd^{\perp}$ and $\hat{\bc}/\| \hat{\bc} \|_2$ are unit $\ell_2$-norm vectors. Thus,
\begin{eqnarray}
\begin{aligned}
\sum_{\bd_i \in \bB \atop i \in \calI^c_0} \epsilon_i (\bd^{\perp})^T  \bd_i^{\hat{\bc}^{\perp}} \le \left| \sum_{\bd_i \in \bB \atop i \in \calI^c_0}  \epsilon_i {\bd^{\perp}}^T \bd_i \right| + \left| \sum_{\bd_i \in \bB } \epsilon_i \frac{ \hat{\bc}^T \bd_i}{\| \hat{\bc} \|_2}  \right|
\end{aligned}
\label{eq:baadz}
\end{eqnarray}
According to Lemma~\ref{lm:projectranodm}
\begin{eqnarray}
 \max \left( \{ \bb_i^T \bd^{\perp} \}_{b_i \neq \bd} \right) < \sqrt{\frac{ c_{\delta}^{''} }{M_1 }}
\label{eq:komaki1}
\end{eqnarray}
with probability at least $1 - \delta$ for all $\{ \bd = \bb_i \}_{i=1}^{n_o}$
where $\sqrt{c_{\delta}^{''}} = 3 \max \left(1 ,  \sqrt{\frac{8 \pi M_1   }{M  - 1}} , \sqrt{\frac{16 M_1  \log n_o/\delta }{M_1 - 1}} \right) $.
Suppose (\ref{eq:komaki1}) is true.    Lemma~\ref{lm:ramche} is used to bound the first part of the RHS of (\ref{eq:baadz}) as
\begin{eqnarray}
\mathbb{P} \left[ \left| \sum_{\bd_i \in \bB \atop i \in \calI^c_0}  \epsilon_i {\bd^{\perp}}^T \bd_i \right| > \sqrt{ (n_o - n_z) \frac{ c_{\delta}^{''} \log n_o/\delta}{M_1 }} \right] < \delta/n_o\:.
\label{eq:shr2}
\end{eqnarray}

In order to bound the second part of the RHS of (\ref{eq:baadz}), first we define vector $\ba$ such that $\ba(i) = \frac{ \hat{\bc}^T \bd_i}{\| \hat{\bc} \|_2}$.
Note that
\begin{eqnarray*}
\| \ba \|_2^2 =   \sum_{\bd_i \in \bB } | \frac{1}{\|\hat{\bc}\|_2} \hat{\bc}^T \bd_i|^2 < \underset{\| \delta \| = 1}{\sup} \sum_{\bd_i \in \bB} |\delta^T \bd_i|^2 \:.
\end{eqnarray*}
Thus,
according to Lemma~\ref{lm:mylemma},
\begin{eqnarray}
\mathbb{P} \left[ \| \ba \|_2 > \sqrt{\frac{n_o}{M_1} + \eta_{\delta}} \right] \le \delta
\label{eq:compreh_2q}
\end{eqnarray}
where $ \eta_{\delta} = \max \left( \frac{4}{3} \log \frac{2M_1}{\delta} , \sqrt{4 \frac{n_o}{M_1} \log \frac{2 M_1}{\delta}} \right) $.
Note that (\ref{eq:compreh_2q}) is true for all $\{\hat{\bc} = \bc_i^{*} \}_{i=1}^{n_o}$.

If (\ref{eq:compreh_2q}) is true,  Lemma~\ref{lm:ramche} can be used to conclude that
\begin{eqnarray}
\mathbb{P} \left[\left| \sum_{\bd_i \in \bB } \epsilon_i \frac{ \hat{\bc}^T \bd_i}{\| \hat{\bc} \|_2}  \right|  > \sqrt{  \left( \frac{n_o}{M_1 } + \eta_{\delta} \right)\log \frac{n_o}{\delta}  } \right] \le  \delta/n_o \: .
\label{eq:shr3}
\end{eqnarray}
\noindent
If (\ref{eq:komaki1}) is true, then
\begin{eqnarray}
 \sum_{\bd_i \in \bB \atop i \in \calI_0}  \left|  \bd_i^T \bd^{\perp}  \right| < n_z \sqrt{\frac{c_{\delta}''}{M_1}}
\label{eq:shr4}
\end{eqnarray}

\noindent
If (\ref{eq:new_general_bound}) is true, then
\begin{eqnarray}
 \frac{\| \bd^T \bU \|_2}{ \|\bd^T \bU^{\perp}\|_2 } < \sqrt{\frac{M_1 \: r \: c_{\delta}}{M_1 (M_1 - c_{\delta} r)}}
\label{eq:cof_boundnn}
\end{eqnarray}

\noindent
According to (\ref{eq:shr1}),(\ref{eq:shr2}),(\ref{eq:shr3}), (\ref{eq:shr4}), and (\ref{eq:cof_boundnn}), if (\ref{eq:second_goy}) is satisfied,  the second inequality of (\ref{suf:mainone}) is satisfied with probability at least $1-4\delta$ for all $\{\bd = \bb_i \}_{i=1}^{n_o}$.

\smallbreak
\noindent
\textbf{Proof of Lemma~\ref{lm:intermediate_last}} \\
\noindent
Suppose $\bd$ in (\ref{opt:asli1}) is an inlier. According to the linear constraint of (\ref{opt:asli1}), $\| \bU^T \bc^{*} \|_2 \ge 1$ where $\bc^{*}$ is the optimal point of (\ref{opt:asli1}).
Therefore,
\begin{eqnarray}
\begin{aligned}
 \| \bD^T \bc^{*} \|_1 =& \|\bA^T \bc^{*} \|_1 + \|\bB^T \bc^{*} \|_1 \\
\ge& \underset{\delta \in \calU  \atop \| \delta \|_2 = 1}{\inf} \| \delta^T \bA \|_1 +  \underset{ \| \delta \|_2 = 1}{\inf}  \| \delta^T \bB \|_1
\end{aligned}
\label{eq:new_tozih}
\end{eqnarray}
Lemma~\ref{lm:perm_negative} can be used to bound  the RHS of (\ref{eq:new_tozih}). Therefore,
\begin{eqnarray}
\begin{aligned}
&  \mathbb{P} \Big[ \|\bD^T \bc^{*} \|_1 <  \sqrt{\frac{2}{\pi}} \frac{n_i}{\sqrt{r}} - 2\sqrt{n_i} - \sqrt{\frac{2 n_i \log \frac{1}{\delta}}{r -1 }} \\
& \quad + \sqrt{\frac{2}{\pi}} \frac{n_o}{\sqrt{M_1}} - 2\sqrt{n_o} - \sqrt{\frac{2 n_o \log \frac{1}{\delta}}{M -1 }} \Big]  < 2\delta  \:.
\end{aligned}
\label{eq:hdgdf22}
\end{eqnarray}
Note that (\ref{eq:hdgdf22}) holds for all $\{\bd = \bd_i \: : \: \bd_i \in \bA \}$.

%\noindent
Next we bound the inverse of the Innovation Value corresponding to an outlying column. Suppose $\bd$ is an outlier. If the sufficient conditions of Lemma~\ref{lm:suffic_random} are satisfied, $\bc^{*}$ lies in $\calU^{\perp}$ with high probability. If $\bc^{*} \in \calU^{\perp}$, ${\bc^{*}}^T \bD = {\bc^{*}}^T \bB$. In addition,
\begin{eqnarray*}
\| \bB^T \bc^{*} \| \le \|\bc^{*} \| \underset{ \| \delta \|_2 = 1}{\sup} \| \delta^T \bB \|_1 \:.
\end{eqnarray*}
Lemma~\ref{lm:perm_positive} is used to establish the following bound on the inverse of the Innovation Value corresponding to an outlying column
\begin{eqnarray}
\begin{aligned}
&  \mathbb{P} \Bigg[ \|{\bc^*}^T \bD \|_1 >   \frac{ \chi n_o}{\sqrt{M_1}} + 2\chi\sqrt{n_o} + \sqrt{\frac{2 \chi n_o \log \frac{1}{\delta}}{M_1 -1 }} \Bigg] \\
&\quad \quad \quad \quad \quad \quad < \delta  \:.
\end{aligned}
\label{eq:new_editf2}
\end{eqnarray}
Note that (\ref{eq:new_editf2}) holds for all $\{\bd = \bd_i \: : \: \bd_i \in \bB \}$.
Thus if the inequality (\ref{eq:akahar}) is satisfied, (\ref{cond:Innovation_condition})  holds with probability at least $1 - 3 \delta$.

\smallbreak
\noindent
\textbf{Proof of Lemma~\ref{lm:orth_clust}}\\
According to the analysis presented in the proof of Lemma~\ref{lm:main_intr}, it is enough to guarantee that (\ref{cond_lemman2}) holds.
Since $\eta < |\bq^T \bq^{\perp} |$, the set $\calI_0 = \{ i \in [M_2] : \hat{\bc}^T \bd_i = 0,\: \bd_i \in \bB \}$ is empty.
For the first inequality of (\ref{cond_lemman2}), we need to bound $\underset{\delta \in \calU  \atop \| \delta \| = 1}{\sup} \delta^T \bo$ which can be expanded as follows
\begin{eqnarray}\notag
\begin{aligned}
\underset{\delta \in \calU  \atop \| \delta \| = 1}{\sup} \delta^T \bo \le&  \frac{1}{\sqrt{1 + \eta^2}} \underset{\delta \in \calU  \atop \| \delta \| = 1}{\sup} \left( n_o  \delta^T\bq + \eta \sum_{i=1}^{n_o} \bv_i^T \delta  \right)\\ <
& \quad \frac{1}{\sqrt{1 + \eta^2}} \left( n_o \|\bU^T \bq \|_2 + \eta \underset{\delta \in \calU  \atop \| \delta \| = 1}{\sup} \sum_{i=1}^{n_o} \bv_i^T \delta \right)
\end{aligned}
%\label{eq:bedebiad1}
\end{eqnarray}

The distribution of $\underset{\delta \in \calU  \atop \| \delta \| = 1}{\sup} \sum_{i=1}^{n_o} \bv_i^T \delta$ is equivalent to the distribution of $\underset{\delta \in \calU  \atop \| \delta \| = 1}{\sup} \sum_{i=1}^{n_o} \epsilon_i \bv_i^T \delta$ where $\{ \epsilon_i \}_{i=1}^{n_o}$ are independent Rademacher random variables. In addition, $\bv_i^T \delta \le \| \bU^T \delta \|_2$.
If (\ref{eq:new_general_bound}) is true, we can use Lemma~\ref{lm:ramche} to establish the following bound
\begin{eqnarray}
\begin{aligned}
& \mathbb{P} \Bigg[ \underset{\delta \in \calU  \atop \| \delta \| = 1}{\sup} \delta^T \bo > \\
& \frac{1}{\sqrt{1+\eta^2}} \left( n_o \|\bU^T \bq \|_2 + \eta  \sqrt{\frac{n_o r c_{\delta} \log n_o/\delta}{M_1} } \right) \Bigg]   <   \delta / n_o \:.
\end{aligned}
\label{eq:tabe1}
\end{eqnarray}

\noindent
In order to derive the sufficient conditions to guarantee that the second inequality of (\ref{cond_lemman2}) holds, we need to bound $\underset{\delta_o \in \: \calU^{\perp} 	\cap \calQ  \atop \delta_o^T \bd = 1}{\sup} \: \delta_o^T  \bo$
which is equal to $ \frac{1}{\bd^T \bq^{\perp}} (\bq^{\perp})^T  \bo$.
According to Lemma~\ref{lm:projectranodm},
\begin{eqnarray}
 \max \left( \{ \bv_i^T \bq^{\perp} \}_{i=1}^{n_o} \right) < \sqrt{\frac{ c_{\delta}^{''} }{M_1 }}
\label{eq:komaki155}
\end{eqnarray}
with probability at least $1 - \delta$ where $\sqrt{c_{\delta}''} = 3 \max \left(1 ,  \sqrt{\frac{8  M_1 \pi }{M_1 - 1}} , \sqrt{\frac{8 M_1 \log n_0/\delta }{M_1 - 1}} \right)$.
If (\ref{eq:komaki155}) is true, $ \frac{1}{\bd^T \bq^{\perp}} (\bq^{\perp})^T  \bo$ can be upper-bounded as
\begin{eqnarray}
\begin{aligned}
& \mathbb{P} \Bigg[ \underset{\delta_o \in \: \calU^{\perp} 	\cap \calD  \atop \delta_o^T \bd = 1}{\sup} \: \delta_o^T  \bo  > \frac{1}{| \bd^T \bq^{\perp} | \sqrt{1+\eta^2}} \Big( n_o |\bq^T \bq^{\perp} | + \\
& \quad \quad\quad\quad\quad\quad\quad \quad\quad \eta  \sqrt{\frac{n_o  c_{\delta}'' \log n_o/\delta}{M_1} } \Big) \Bigg]  <  \delta/n_o \:.
\end{aligned}
\label{eq:tabe2}
\end{eqnarray}

The LHS of the first inequality of (\ref{cond_lemman2}) is bounded similar to (\ref{eq:suf_1_1}). The LHS of the second inequality of (\ref{cond_lemman2}) can be bounded similar to (\ref{linear_cond2}). Therefore, according to (\ref{eq:tabe1}),(\ref{eq:tabe2}),(\ref{eq:suf_1_1}),and (\ref{linear_cond2}) if (\ref{eq:sf11}) and (\ref{eq:sf112}) are satisfied, then the inequalities of (\ref{cond_lemman2}) hold  for all $\{\bd = \bd_i \: : \: \bd_i \in \bB\}$ with probability at least $1 - 5\delta$, i.e., the optimal point of (\ref{opt:withspan}) is equal to  $\frac{1}{\bd^T \bq^{\perp}} \bq^{\perp}  $ with probability at least $1 - 5\delta$ for all $\{\bd = \bd_i \: : \: \bd_i \in \bB\}$.

\smallbreak
\noindent
\textbf{Proof of Lemma~\ref{lm:finallle}}\\
 First, we derive a lower-bound for  $\| \bc^{*} \bD \|_1$
when $\bd$ is an inlier.
According to the first linear constraint of (\ref{opt:withspan}), $\| \bc^{*} \|_2 \ge 1$. In addition, $\| {\bc^{*}}^T \bD \|_1 \ge \| {\bc^{*}}^T \bA \|_1$ and $$\| {\bc^{*}}^T \bA \|_1 \ge \underset{ \| \delta\|_2 = 1 \atop \delta \in \calU}{\inf} \| \delta^T \bA\|_1 \:.$$ We use Lemma~\ref{lm:perm_negative} to conclude that if $\bd \in \calA$
\begin{eqnarray}
\begin{aligned}
&  \mathbb{P} \Big[ \| {\bc^*}^T \bD \|_1 <  \sqrt{\frac{2}{\pi}} \frac{n_i}{\sqrt{r}} - 2\sqrt{n_i} - \sqrt{\frac{2 n_i \log \frac{1}{\delta}}{r -1 }}  \Big]  < 1 / \delta   \:.
\end{aligned}
\label{eq:vhgf}
\end{eqnarray}
Note that (\ref{eq:vhgf}) holds for all the inliers with probability at least $1 - \delta$.

If $\bd$ is an outlier and
 the sufficient conditions of Lemma~\ref{lm:orth_clust} are satisfied, $\bc^{*} \in \calU^{\perp} 	\cap \calQ $ with high probability. If $\bc^{*} \in \calU^{\perp} 	\cap \calQ $, then $\| {\bc^{*}}^T \bD \|_1 = \| {\bc^{*}}^T \bB\|_1$.
 We can expand $ \|{\bc^{*}}^T \bB \|_1$ as
\begin{eqnarray*}
\begin{aligned}
 \| {\bc^{*}}^T \bB \|_1 =& \frac{\| \bc^{*} \|_2}{\sqrt{1+\eta^2}} \sum_{i=1}^{n_o} \left| \bq^T \bq^{\perp} + \eta \bv_i^T \bq^{\perp} \right| \\
 \le&
  \frac{\| \bc^{*} \|_2}{\sqrt{1+\eta^2}} \left( n_o  | \bq^T \bq^{\perp} | + \eta \sum_{i=1}^{n_o} \left|  \bv_i^T \bq^{\perp} \right| \right)
\end{aligned}
\end{eqnarray*}
Thus, we can use Lemma~\ref{lm:projectranodm} to conclude that
\begin{eqnarray*}
\begin{aligned}
& \mathbb{P} \Bigg[
 \| {\bc^{*}}^T \bB \|_1 \ge \\
 & \frac{\| \bc^{*} \|_2}{\sqrt{1+\eta^2}} \left( n_o |\bq^T \bq^{\perp}| + \eta n_o \sqrt{\frac{  c_{\delta}'' \log n_o/\delta}{M_1} }  \right) \Bigg]  \le \delta
\end{aligned}
\end{eqnarray*}
where $\sqrt{c_{\delta}''} = 3 \max \left(1 ,  \sqrt{\frac{8  M_1 \pi }{M_1 - 1}} , \sqrt{\frac{8 M_1 \log n_0/\delta }{M_1 - 1}} \right)$. Thus, if (\ref{eq:tabozha}) is satisfied  the outliers are guaranteed to have greater Innovation Values with probability at least $1 - 2\delta$.

\smallbreak
\noindent
\textbf{Proof of Lemma~\ref{lm:orh_sb_clus}}\\
Lemma~\ref{lm:main_intr} does not make any assumption about the distribution of the outliers. Accordingly, we only need to guarantee that the sufficient conditions of Lemma (\ref{lm:main_intr}) are satisfied.
First we bound the RHS of the first inequality in (\ref{suf:mainone}). Since the columns of $\bB$ lie in $\calU_o$, the $\ell_2$-norm of the projection of the columns of $\bB$ into $\calU$ is bounded by $\| \bU^T \bU_o \|$. Accordingly,
\begin{eqnarray}\notag
\underset{\delta \in \calU  \atop \| \delta \| = 1}{\sup} \sum_{\bd_i \in \bB \atop i \in \calI_0}  \left| \delta^T \bd_i \right| \leq n_z \| \bU^T \bU_o \|
%\label{eq:suf_1_2}
\end{eqnarray}
where $n_z = |\calI_0|$ and $\calI_0 = \{ i \in [n_o] : {\bc^{*}}^T \bb_i = 0\}$.

\noindent
Similar to (\ref{eq:oooo}),  $\bo$  can be expanded as
\begin{eqnarray*}
\begin{aligned}
\bo = \sum_{\bd_i \in \bB} \frac{|\hat{\bc}^T \bd_i|}{\| \hat{\bc} \|_2^2} \hat{\bc} + \sum_{\bd_i \in \bB} \sgn (\hat{\bc}^T \bd_i) \:  \bd_i^{\hat{\bc}^{\perp}} \:.
\end{aligned}
\end{eqnarray*}

The distribution of $\bo$ is equivalent to the distribution of $ \bo^{'} = \sum_{\bd_i \in \bB \atop i \in \calI^c_0} \frac{|\hat{\bc}^T \bd_i|}{\| \hat{\bc} \|_2^2} \hat{\bc} + \sum_{\bd_i \in \bB \atop i \in \calI^c_0}   \:   \epsilon_i \bd_i^{\hat{\bc}^{\perp}}$, where $\{ \epsilon_i\}_i$   are independent Rademacher random variables. The vector $\bd_i^{\hat{\bc}^{\perp}}$ is defined similar to (\ref{eq:dcdo}). Since $\hat{\bc}$ is orthogonal to $\calU$
\begin{eqnarray}
\begin{aligned}
\underset{\delta \in \calU  \atop \| \delta \| = 1}{\sup} \delta^T \bo^{'} =& \underset{\delta \in \calU  \atop \| \delta \| = 1}{\sup} \left( \sum_{\bd_i \in \bB \atop i \in \calI^c_0} \epsilon_i \: \delta^T  \bd_i^{\hat{\bc}^{\perp}} \right)\\
=
&\underset{\delta \in \calU  \atop \| \delta \| = 1}{\sup} \left( \sum_{\bd_i \in \bB \atop i \in \calI^c_0} \epsilon_i \: \delta^T  \bd_i \right) \: .
\end{aligned}
\label{eq:ayasa}
\end{eqnarray}

\noindent
Since $\delta \in \calU$, $\delta^T \bd_i \le \| \bU^T \bU_o \|$. Accordingly, we use Lemma~\ref{lm:ramche} to bound (\ref{eq:ayasa})
\begin{eqnarray}\notag
\begin{aligned}
 & \mathbb{P} \Bigg[ \underset{\delta \in \calU  \atop \| \delta \| = 1}{\sup} \left( \sum_{\bd_i \in \bB \atop i \in \calI^c_0} \epsilon_i \: \delta^T  \bd_i \right) >  \\
 & \quad\quad\quad\quad \| \bU^T \bU_o \| \sqrt{(n_o - n_z)  \log n_o/\delta } \Bigg]  \le \delta/n_o \: .
\end{aligned}
%\label{eq:ayasa2}
\end{eqnarray}

The LHS of the first inequality of (\ref{suf:mainone}) is bounded similar to (\ref{eq:suf_1_1}). Thus, if (\ref{eq:suff_1_orth_dep}) is satisfied, the first inequality of (\ref{suf:mainone}) holds for all $\{ \bd= \bd_i \: : \: \bd_i \in \bB \}$ with probability at least $1 - 2\delta$.

In order to guarantee that the second inequality of (\ref{suf:mainone}) holds, first we bound the absolute value of  ${\bo^{'}}^T \bd^{\perp}$. Similar to (\ref{eq:odperp})
\begin{eqnarray}
\begin{aligned}
{\bo^{'}}^T \bd^{\perp} =& \sum_{\bd_i \in \bB \atop i \in \calI^c_0} \frac{|\hat{\bc}^T \bd_i|}{\| \hat{\bc} \|_2^2} \hat{\bc}^T \bd^{\perp} + \sum_{\bd_i \in \bB \atop i \in \calI^c_0} \epsilon_i \: (\bd^{\perp})^T  \bd_i^{\hat{\bc}^{\perp}}\\
\le&\frac{1}{\|\hat{\bc}\|_2} \sum_{\bd_i \in \bB } {|\hat{\bc}^T \bd_i|}  + \sum_{\bd_i \in \bB \atop i \in \calI^c_0} \epsilon_i (\bd^{\perp})^T  \bd_i^{\hat{\bc}^{\perp}}
\end{aligned}
\label{eq:odperp2}
\end{eqnarray}

Define $\hat{\bc}_o =  \frac{ \bU_o \bU_o^T \hat{\bc}}{\| \bU_o \bU_o^T \hat{\bc} \|_2}$. In addition, $ \frac{1}{\| \hat{c}\|_2} \| \bU_o^T \hat{\bc} \|_2 \le \| \bU_o^T \bU^{\perp} \| $. Accordingly,
\begin{eqnarray*}
\frac{1}{\|\hat{\bc}\|_2} \sum_{\bd_i \in \bB } {|\hat{\bc}^T \bd_i|} \le \|  \bU_o^T \bU^{\perp} \|  \sum_{\bd_i \in \bB } | \hat{\bc}_o^T \bd_i|\:.
\end{eqnarray*}
Moreover,
\begin{eqnarray}
\|  \bU_o^T \bU^{\perp} \|  \sum_{\bd_i \in \bB } | \hat{\bc}_o^T \bd_i| \le \|  \bU_o^T \bU^{\perp} \| \underset{\delta \in \calU_{o}  \atop \| \delta \| = 1}{\sup} \sum_{\bd_i \in \bB } |\delta^T \bd_i|\:.
\label{eq:technieinperp}
\end{eqnarray}
Accordingly, we can use Lemma~\ref{lm:perm_positive} to bound the first component of the RHS of (\ref{eq:odperp2})   as follows
\begin{eqnarray}
\begin{aligned}
&\mathbb{P} \Bigg[ \frac{1}{\|\hat{\bc}\|_2} \sum_{\bd_i \in \bB } {|\hat{\bc}^T \bd_i|}  > \|\bU_o^T \bU^{\perp} \| \Bigg( \frac{n_o }{\sqrt{ \textcolor{black}{ r_o}}} + 2\sqrt{n_o } +  \\
& \quad \quad \sqrt{\frac{2 n_o \log \frac{1}{\delta}}{r_o -1 }} \Bigg) \Bigg] <\delta \:.
\end{aligned}
\label{eq:shr12}
\end{eqnarray}
Note that (\ref{eq:shr12}) is true for all $\{\bd = \bd_i \: : \: \bd_i \in \bB \}$. Similar to (\ref{eq:nnw1}) and (\ref{eq:baadz}), the second component in the RHS of (\ref{eq:odperp2}) can be expanded as follows
\begin{eqnarray}\notag
\begin{aligned}
\sum_{\bd_i \in \bB \atop i \in \calI^c_0} \epsilon_i (\bd^{\perp})^T  \bd_i^{\hat{\bc}^{\perp}} \le \left| \sum_{\bd_i \in \bB \atop i \in \calI^c_0}  \epsilon_i {\bd^{\perp}}^T \bd_i \right| + \left| \sum_{\bd_i \in \bB } \epsilon_i \frac{ \hat{\bc}^T \bd_i}{\| \hat{\bc} \|_2}  \right| \: .
\end{aligned}
%\label{eq:baadz2}
\end{eqnarray}

Define  vector $\ba$ such that $\ba(i) = \frac{\hat{\bc}^T \bd_i}{\| \hat{\bc} \|_2 }$. According to Lemma~\ref{lm:mylemma},
\begin{eqnarray}
\mathbb{P} \left[ \| \ba \|_2 >  \| {\bU^{\perp}}^T \bU_o \| \sqrt{ \frac{n_o}{r_o} + \eta_{\delta}^{'} } \: \right] \le \delta
\label{eq:condonann}
\end{eqnarray}
where $\eta_{\delta}^{'} = \max \left(\frac{4}{3} \log 2 r_o /\delta \: , \: \sqrt{4 \frac{n_o}{r_o} \log \frac{2 r_d}{\delta}} \right) $. If (\ref{eq:condonann})  is true,   Lemma~\ref{lm:ramche} can be used to conclude that
\begin{eqnarray}
\begin{aligned}
& \mathbb{P} \left[\left| \sum_{\bd_i \in \bB } \epsilon_i \frac{ \hat{\bc}^T \bd_i}{\| \hat{\bc} \|_2}  \right|  > \| {\bU^{\perp}}^T \bU_o \| \sqrt{  \left( \frac{n_o}{r_o} + \eta_{\delta}^{'} \right)\log \frac{n_o}{\delta}  } \right] \\
& \quad\quad\quad\quad\quad \le  \delta/n_o \: .
\label{eq:shr3h}
\end{aligned}
\end{eqnarray}

Similar to $\hat{\bc} / \| \hat{\bc} \|$, $\bd^{\perp}$ lies in $\calU^{\perp}$ and $\| \bd^{\perp} \| = 1$. Thus, we can bound the second component of the RHS of (\ref{eq:odperp2}) similarly
\begin{eqnarray}
\begin{aligned}
& \mathbb{P} \left[ \left| \sum_{\bd_i \in \bB \atop i \in \calI^c_0} \epsilon_i (\bd^{\perp})^T  \bd_i^{\hat{\bc}^{\perp}} \right| > \| {\bU^{\perp}}^T \bU_o \| \sqrt{  \left( \frac{n_o}{r_o} + \eta_{\delta}^{'} \right)\log \frac{n_o}{\delta}  } \right] \\
& \quad\quad\quad\quad\quad \le  \delta/n_o \: .
\label{eq:shr3hh}
\end{aligned}
\end{eqnarray}
The last component of the RHS of the second inequality of (\ref{suf:mainone}) can be bounded as
\begin{eqnarray}
\sum_{\bd_i \in \bB \atop i \in \calI_0}  \left|  \bd_i^T \bd^{\perp}  \right| \le n_z \| {\bU^{\perp}}^T \bU_o \| \:.
\label{eq:final_tingd1}
\end{eqnarray}

According to the definition of $\xi$, the coefficient $$\frac{ \| \bd^T \bU \|_2}{\sqrt{1 - \|\bd^T \bU \|_2^2}} = \frac{ \| \bd^T \bU \|_2}{\| \bd^T \bU^{\perp} \|}$$ can be bounded as
\begin{eqnarray}
\frac{ \| \bd^T \bU \|_2}{\| \bd^T \bU^{\perp} \|} \le \frac{\| \bU^T \bU_o \|}{\xi \| {\bU^{\perp}}^T \bU_o \|}.
\label{eq:coff_bound1}
\end{eqnarray}

Using (\ref{eq:shr12}), (\ref{eq:shr3h}), (\ref{eq:shr3hh}), (\ref{eq:final_tingd1}), and (\ref{eq:coff_bound1}) we can establish an upper-bound on the RHS of the second inequality of (\ref{suf:mainone}).  Thus if (\ref{eq:suff_2_orth_dep}) is satisfied, the second inequality of (\ref{suf:mainone}) holds with probability at least  $1 - 4 \delta$ for all $\{ \bd=\bd_i \: : \: \bd_i \in \bB \}$.

\smallbreak
\noindent
\textbf{Proof of Lemma~\ref{lm:values_sub_lin_dep}} \\
If the sufficient conditions of Lemma (\ref{lm:orh_sb_clus}) are satisfied and if $\bd \in \bB$, then $\bc^{*}$ lies in $\calU^{\perp}$ with high probability. If $\bc^{*} \in \calU^{\perp}$, then $\| \bD^T \bc^{*}\|_1 = \| \bB^T \bc^{*} \|_1$.
Assuming that $\bc^{*} \in \calU^{\perp}$, we use Lemma~\ref{lm:perm_positive} and inequality (\ref{eq:technieinperp}) to conclude that
\begin{eqnarray*}
\begin{aligned}
&  \mathbb{P} \Bigg[ \|{\bc^*}^T \bD \|_1 >  \|\bU_o^T \bU^{\perp} \|_2 \Bigg(  \frac{ \chi n_o}{\sqrt{r_o}} + 2\sqrt{\chi n_o} + \\
&  \quad \quad \quad \sqrt{ \frac{2 \chi n_o \log \frac{1}{\delta}}{r_o }} \Bigg) \Bigg] < \delta  \:.
\end{aligned}
\end{eqnarray*}

If $\bd \in \bA$, the linear constraint of (\ref{opt:asli1}) ensures that $\| \bU^T \bc^{*} \|_2 \ge 1$.
In addition, $\| \bD^T \bc^{*}\|_1 \ge \| \bA^T \bc^{*} \|_1$.
Accordingly, when $\bd \in \bA$, we can use Lemma~\ref{lm:perm_negative} to derive the following bound on $\| \bD^T \bc^{*} \|_1$
\begin{eqnarray}\notag
\begin{aligned}
&  \mathbb{P} \left[ \|{\bc^*}^T \bD \|_1 <  \sqrt{\frac{2}{\pi}} \frac{n_i}{\sqrt{r}} - 2\sqrt{n_i} - \sqrt{\frac{2 n_i \log \frac{1}{\delta}}{r -1 }}  \right] < \delta \: .
\end{aligned}
%\label{eq:bound_inn1}
\end{eqnarray}
Thus, if (\ref{eq:valueinlemmasub}) is satisfied,  the Innovation Values corresponding to the outliers are lager than the corresponding values for the inliers.

\end{document}